\documentclass{article}

\usepackage[utf8]{inputenc}

\usepackage{cite}
\usepackage{comment}
\usepackage[square,sort,comma,numbers]{natbib}
\usepackage{graphicx}
\usepackage{color}
\usepackage[dvipsnames]{xcolor}
\usepackage{amsmath}
\usepackage{amsfonts}
\usepackage{amssymb}
\usepackage{enumitem}
\usepackage{mathtools}
\usepackage{soul}
\usepackage{setspace}
\usepackage{url}
\usepackage{tabularx} % Tables
\usepackage{array}
\usepackage{multirow}
\usepackage{colortbl}
\usepackage{color}
\usepackage{lscape}
\definecolor{my_blue_lite}{rgb}{0.886,0.921,0.992}
\definecolor{my_green_lite}{rgb}{0.859,0.988,0.89}
\definecolor{my_purple_lite}{rgb}{0.980,0.859,0.988}
\usepackage{ragged2e} % for nicer table rules
\usepackage{booktabs} % define a new column type for convenience
\newcolumntype{L}[1]{>{\hsize=#1\hsize\RaggedRight} X}

\let\comments\undefined %% the comments will not be compiled 

\ifdefined\comments
    
    \newcommand{\RemoveForShortVersion}[1]{\textcolor{gray}{#1}}
    \newcommand{\Alexis}[1]{\smallskip\noindent{}\textcolor{red}{{\bf A.Bo}: #1}}
    \newcommand{\Vincent}[1]{\smallskip\noindent{}\textcolor{magenta}{{\bf V.Le}: #1}}
    \newcommand{\Youssef}[1]{\smallskip\noindent{}\textcolor{BlueViolet}{{\bf Y.Ac}: #1}}
    \newcommand{\Antoine}[1]{\smallskip\noindent{}\textcolor{BlueGreen}{{\bf A.Co}: #1}}
    \newcommand{\Albert}[1]{\smallskip\noindent{}\textcolor{Salmon}{{\bf A.Bi}: #1}}
    \newcommand{\Georges}[1]{\smallskip\noindent{}\textcolor{Mulberry}{{\bf G.He}: #1}}
    \newcommand{\PierreFrancois}[1]{\smallskip\noindent{}\textcolor{Bittersweet}{{\bf PF.M}: #1}}
    \newcommand{\Joao}[1]{\smallskip\noindent{}\textcolor{YellowGreen}{{\bf J.Ga}: #1}}
\else

    \newcommand{\RemoveForShortVersion}[1]{}
    \newcommand{\Alexis}[1]{}
    \newcommand{\Vincent}[1]{}
    \newcommand{\Youssef}[1]{}
    \newcommand{\Antoine}[1]{}
    \newcommand{\Albert}[1]{}
    \newcommand{\Georges}[1]{}
    \newcommand{\PierreFrancois}[1]{}
    \newcommand{\Joao}[1]{}
\fi

\newcommand{\expectancy}[3]{ \operatornamewithlimits{\mathbb{E}^{\:#3}}_{\:#1} \left [ #2 \right ] } 
\DeclareMathOperator*{\argmin}{arg\,min}
\DeclareMathOperator*{\LL}{\mathcal{L \hspace{-1.8mm} L}}

\begin{document}

\title{Open challenges for Machine Learning based \\ Early Decision-Making research}

% Authors
\author{Alexis Bondu, Youssef Achenchabe, Albert Bifet, Fabrice Cl\'erot, \\ Antoine Cornu\'ejols, Joao Gama, Georges H\'ebrail,\\ Vincent Lemaire, Pierre-François Marteau \Youssef{affiliations??}}

% The paper headers IEEE TRANSACTIONS ON KNOWLEDGE AND DATA ENGINEERING, VOL. 22, NO. 10, OCTOBER 2010
%\markboth{IEEE Transaction on Knowledge and Data Engineering, Vol.~XX No.~XX, Month Year}%
%{Bondu \MakeLowercase{\textit{et al.}}: Open challenges for ML-EDM research}

\maketitle

% Abstract
%\IEEEtitleabstractindextext{% % CHANGE TEMPLATE
\begin{abstract}
More and more applications require \textit{early} decisions, i.e. taken as soon as possible from partially observed data. However, the \textit{later} a decision is made, the more its accuracy tends to improve, since the description of the problem to hand is enriched over time. Such a compromise between the \textit{earliness} and the \textit{accuracy} of decisions has been particularly studied in the field of Early Time Series Classification. This paper introduces a more general problem, called Machine Learning based Early Decision Making (ML-EDM), which consists in optimizing the decision times of models in a wide range of settings where data is collected over time. After defining the ML-EDM problem, ten challenges are identified and proposed to the scientific community to further research in this area. These challenges open important application perspectives, discussed in this paper. 
\end{abstract}

%\begin{IEEEkeywords} % CHANGE TEMPLATE
% Early Decision-Making, Machine Learning, Time Evolving Data, Online Decision-Making, Revocable Decisions. % CHANGE TEMPLATE
% \end{IEEEkeywords}} % CHANGE TEMPLATE


%%%%%%%%%%%%%%%%% Table of content %%%%%%%%%%%%%%%%%%%%

%\renewcommand\contentsname{Table of contents}
%\setcounter{tocdepth}{1}

%\begin{spacing}{0.9}
%\tableofcontents
%\end{spacing}

%-------------------------------------------------------------------
%-------------------------------------------------------------------
\section{Introduction}
%-------------------------------------------------------------------
%-------------------------------------------------------------------

In numerous real situations, we have to make \textit{early} decisions in the absence of \textit{complete knowledge} 
of the problem at hand. 
For example, such decisions are necessary in medicine \citep{mathukia2015modified} when a physician must make a diagnosis, possibly leading to an urgent surgical operation, before having the results of all medical tests. 
In such situations, the issue facing the decision makers is that, most of the time, the longer the decision is delayed, the clearer is the likely outcome (e.g. the critical or not critical state of the patient) but, also, the higher the cost that will be incurred if only because decisions taken earlier allow one to be better prepared. 
We thus seek to make decisions at times that seem to be the best compromises between the \textit{earliness} and the \textit{accuracy} of our decisions. %For instance, when driving, we don't wait until we are sure that there is a traffic jam ahead of us to decide to change our route. 

\smallskip
Similarly in Machine Learning, when the input data is acquired over time, there can be situations with a trade-off between the earliness and accuracy of decisions. 
For instance, this is the case for anomaly detection, predictive maintenance, patient health monitoring, self-driving vehicles (see Section \ref{usecases}). In each case, the decisions are \textit{time-sensitive} (e.g. in an autonomous car, it is critical to detect obstacles on the road as early as possible and at the same time as reliably as possible, in order to plan safe avoidance trajectories if needed). In general, it is assumed that there is a \textit{gain of information} over time, i.e. delaying decisions tends to make them more reliable (e.g. the certainty about the existence or absence of an obstacle on the road becomes more and more accurate as the car gets closer).% to it).

\smallskip
This earliness \textit{vs.} accuracy dilemma is part of many decision making scenarios, and is particularly involved in the problem of Early Classification of Time Series (ECTS). But, as we will see, it takes place in a larger perspective.

%-------------------------------------------------------------------

\subsubsection*{Early Classification of Time Series: a particular case}
%~ 
%
\smallskip
%\noindent
The ECTS problem consists in finding the \textit{optimal time} to trigger the class prediction of an input time series observed over time. 
As successive measurements provide more and more information about the incoming time series, ECTS algorithms aim to optimize online the trade-off between the \textit{earliness} and the \textit{accuracy} of their decisions. 

\smallskip
More formally, the individuals\footnote{The term \textit{individual} refers to any type of statistical unit studied.} considered are time series of \textit{finite length} $T$. 
At testing time, %During the recognition phase, 
the measurements of the incoming time series are received over time, and the history of measurements available at time $t$ is denoted by ${\mathbf x}_t \, = \, \langle {x_1}, \ldots, {x_t} \rangle$.
It is assumed that each time series can be ascribed to some class $y \in {\cal Y}$, and the task is to make a prediction about the class of each incoming time series as early as possible, because a time increasing cost must be paid when the decision is triggered.
In the ECTS problem, a single decision is triggered for each incoming time series, which is irrevocable and final.
An ECTS approach is generally made of two main components: 
(\textit{i}) an \textit{hypothesis}\footnote{An hypothesis is a candidate predictor which approximates the concept $P(y|{\mathbf x}_t)$.} $h \in \mathcal{H}$ capable of predicting the class $y \in \mathbb{Y}$ of the incoming series at any time, such that $h({\mathbf x}_t)=\hat{y}$ with $t \in [1,T]$ ; 
(\textit{ii}) a \textit{triggering strategy} capable of making decisions at the right moments, denoted by $Trigger$.
Both the hypothesis and the triggering strategy are learned in batch mode (i.e. offline), by using a training set made of complete time series with their associated labels. 

%-------------------------------------------------------------------

%\Alexis{Fabrice : décrire les données d'apprentissage - une phrase}

\subsubsection*{A short state of the art on Early Classification of Time Series}
%~
%
\smallskip
%\noindent
This paragraph provides an overview of the ECTS approaches. For a recent and more complete survey, the  reader can refer to \cite{9207873, achenchabe2021MLj}.
The pioneering approaches were based on some form of \textit{confidence} criterion and waited until a predefined threshold is reached before triggering their decisions. For instance, in \cite{parrish2013classifying,hatami2013classifiers,ghalwash2012early}, a classifier is learned for each time step and various stopping rules are used (e.g. threshold on confidence level). In \cite{xing2009early}, such a threshold is indirectly set since the best time step to trigger the decision is estimated by determining the earliest time step for which the predicted label does not change, based on a 1NN classifier. Similarly, \cite{mori2017reliable} proposes a method where the accuracy of a set of probabilistic classifiers is monitored over time, which allows the identification of time steps from whence it seems safe to make predictions. 

\smallskip
Then, more informed approaches appeared which explicitly take into account the cost of delaying the decisions. A notable example is \cite{mori2019early} where the conflict between earliness and accuracy is explicitly addressed. Moreover, instead of setting the trade-off in a single objective optimization criterion as in \cite{mori2017early}, the authors keep it as a multi-objective criterion and to explore the Pareto front of the multiple dominating trade-offs. 

\smallskip
The \textsc{Economy} approach \cite{achenchabe2021MLj, dachraoui2015early} goes one step further by casting the ECTS problem as searching to optimize a loss function which combines the expected cost of misclassification at the time of decision, plus the cost of having delayed the decision thus far. 
This well-founded approach is \textit{non-myopic}, as it is able to anticipate measurements which are not yet visible at decision time by estimating the expected costs for future time steps.
This approach leads to the best performances observed to date, and \cite{achenchabe2021MLj} shows that the non-myopic feature of this approach explains its strong performances through an ablation study.

%-------------------------------------------------------------------

\subsubsection*{Limitations of the ECTS problem}

\smallskip
While ECTS covers a wide range of applications, it does not exhaust all cases where a \textit{Machine Learning} model can be applied on data acquired over time, and where the trade-off between the \textit{earliness} and the \textit{accuracy} of decisions must be optimized. Indeed, ECTS, as defined above, is limited to: 

\begin{itemize}
\item a classification problem ; 
\item an available training set which contains completely and properly labeled time series ;
\item a decision deadline $T$ that is finite, fixed and known ;
\item unique decisions for each incoming time series ;
\item decisions that once made can never be reconsidered ;
\item fixed decision costs which do not depend on the triggering time and the decisions made.
\end{itemize}

\noindent
All of these assumptions might be questioned and point to research issues. 
The purpose of this paper is to propose research directions for extending ECTS toward a more generic problem, that we call  \textit{Machine Learning based Early Decision-Making} (ML-EDM). 

\Alexis{TODO : citer le papier de Kheogs pour dire que c'est une question importante ?}
% The issue is that the definition of the problem itself is intrinsically underspecified [...] a consortium of researchers would be better.

\bigskip
This position paper is organised as follows. 
Section \ref{definition-ML-EDM} first defines the ML-EDM problem, shows how a triggering strategy can be learned, and positions ML-EDM with respect to Reinforcement Learning.
A series of ten challenges is then proposed in order to develop ML-EDM approaches for a wide range of problems. 
Section \ref{decision_costs} specifies the decision costs involved in ML-EDM problems, and explains the origin of these costs. 
Section \ref{learning_tasks} considers a variety of \textit{learning tasks}, and Section \ref{data_types}, a variety of \textit{data types}. 
Section \ref{stream} gives some leads to address the problem of \textit{online} ML-EDM. 
Section \ref{revocable_decision} extends ML-EDM to \textit{revocable decisions}. 
Section \ref{sec_overview} gives an overview on the proposed challenges, and makes a synthesis of long and short term application perspectives.
Then, Section \ref{usecases} provides some examples of \textit{applications} of the ML-EDM techniques. 
At last, Section \ref{conclusion} concludes with perspectives for the development of the ML-EDM field in the coming years.

%-------------------------------------------------------------------
%-------------------------------------------------------------------

\section{Definition of ML-EDM}
\label{definition-ML-EDM}

%-------------------------------------------------------------------
%-------------------------------------------------------------------

This section defines  
what ML-EDM is by 
answering the following questions:\\ {\bf A-} What is an early decision? {\bf B-} How to learn a triggering strategy from training data? {\bf C-} Can a triggering strategy be learned by Reinforcement Learning? 

\bigskip
\noindent
\textbf{Question A - }\textit{What is an early decision?}\\
~\\
Basically, Early Decision Making consists in: (\textit{i}) observing pieces of information over time ; (\textit{ii}) deciding when to make a decision ; and (\textit{iii}) making the decision itself. 
In the following, increasingly complex decision-making problems are considered in order to progressively lead to a general definition of ML-EDM.

\medskip
Two types of problems can be distinguished \cite{hansson1994}. 
Decision-making \textit{under ignorance} refers to a category of problems where the set of possible outcomes is known, but no information about their probabilities is available. 
By contrast, decision-making \textit{under uncertainty} deals with problems where the probabilities of the possible outcomes are known, or partially known. 
%This important distinction determines the prior knowledge on which decisions are based. 

\medskip
\noindent
{\bf \textit{Optimal Stopping Problem}} \cite{shepp1969} is a canonical case of interest, where the decision to make is simply to stop receiving new pieces of information. 
More formally, $\{X_i\}$ is a sequence of random variables observed successively, whose joint distribution is known. Let $\{r_i\}$ be a sequence of reward functions, such that $r_i$ is a function of the observed values $x_1 , \dots , x_i$. The objective is to maximize the reward, deciding after observing the value of the random variable $X_i$, either to stop and accept the reward $r_i$, or to observe the value of the next random variable $X_{i+1}$. 
A number of optimal stopping problems have been extensively studied in the literature, such as:

\begin{itemize}
    \item The \textit{Shepp's urn} \cite{shepp1969} which is filled with a known number of \$1 bills, and a known number of anti-bills of -\$1. Here, the reward is the sum of the bills gathered until the end of the game. The objective is to maximize our payoff by stopping to draw objects in this urn at the best time. 
    \item The \textit{secretary problem} \cite{ferguson1989} consists in selecting the largest possible value (which is unknown), among a sequence of values of known size observed in a uniform random order. At each step the choice is, either to stop and keep the last observed value, or to continue. 
\end{itemize}

These two problems involve decision making \textit{under uncertainty}, since the system under study is perfectly known and the probability of the possible outcomes can be estimated. For instance, in the Shepp's urn the probability of getting a bill or an anti-bill in the next draw is available, since the content of the urn is known at any time.
In the secretary problem, 
the rank of the last value among the previously observed values approximates the rank in the entire set of values, since the observed values constitute a uniform sample of all values.

\medskip
As in Early Decision Making problem, Shepp's urn and the secretary problem imply a trade-off between \textit{early} and \textit{accurate} decisions. 

\medskip
On the one hand, there is a \textit{time pressure} which pushes to trigger early  decisions. In a Shepp's urn, the number of objects is finite and if all of them are drawn, our payoff is bad, i.e. equal to the number of bills minus the number of anti-bills. In the secretary problem, the number of values is known. The more values are observed, the less future opportunity remains to select a high value.

\medskip
On the other hand, there is a \textit{gain of information} (about what's left in the urn) over time which tends to delay the decisions. In the Shepp's urn problem, the sample of already drawn objects grows over time, which provides useful information to be compared to the known quantities of bills and anti-bills. For the secretary's problem, the sample of already drawn values grows over time, and the last observed value can be compared to this sample. 

\medskip
From here on, the decision-making problems presented in the following are part of \textit{supervised learning}. A set of labeled examples, which takes different forms depending on the problem, is assumed to be available.

\medskip
\noindent
{\bf \textit{The ECTS problem}} can be considered as a particular instance of optimal stopping, where the decision to be made consists in: (\textit{i}) stopping receiving new measurements ; and (\textit{ii}) predicting the class of the incoming time series. 
The hypothesis $h \in \mathcal{H}$ is assumed to be available, allowing to predict the class $y \in \mathbb{Y}$ of the incoming series at any time, such that $h({\mathbf x}_t)=\hat{y}$.   
In this case, the reward function $r({\mathbf x}_t,t,\hat{y},y)$ depends on the observed measurements ${\mathbf x}_t \, = \, \langle {x_1}, \ldots, {x_t} \rangle$ ; the decision time $t$ ; the predicted class $\hat{y}$ ; and the true class $y$. The following loss function can be defined:  
\begin{equation}
    \mathcal{L}(h({\mathbf x}_t), t, y) = \mathcal{L}_{prediction} \left( h(\mathbf{x}_t),y \right) \; +  \mathcal{L}_{delay}(t) 
\label{eq:loss-function-ECTS}    
\end{equation}

where $\mathcal{L}_{prediction}(.)$ is the cost of making a potentially bad prediction which can be expressed as a cost matrix, and $\mathcal{L}_{delay}(t)$ is a monotonically increasing function of $t$ representing the cost of delaying the decision until $t$\footnote{Note that the delay cost $\mathcal{L}_{delay}(t)$ could depend on the class $y$ of the time series (see Section \ref{decision_costs}). For instance, in the emergency department in a hospital, the cost of delaying a decision when there is internal bleeding is not the same as the one in case of gastroenteritis, where the early symptoms could look the same. 
Here, for reasons of readability, we make $\mathcal{L}_{delay}$ depend only on $t$. 
%As long as $\mathcal{L}_{delay}(t, y)$ can be estimated $\forall{(t,y)}$, the optimization problem can be solved with the same technique.
}. 
%
%\medskip
The best decision time $t^*$ is given by the optimal triggering strategy $Trigger^*$ defined as:

\begin{equation}
\begin{split}
&Trigger^*\left ( h({\mathbf x}_t) \right )  =  \\
&\left\{
    \begin{array}{ll}
        1 & \mbox{if } t = t^* = \argmin_{t \in [1, T]} \mathcal{L}(h({\mathbf x}_t), t, y) \mbox{ or } t=T\\
        0 & \mbox{otherwise}
    \end{array}
\right.
\end{split}
\label{eq:trigger_function}
\end{equation} 
where the decision is forced at $t = T$ if it was not taken before.

\medskip
Here, the trade-off between \textit{early} and \textit{accurate} decisions takes the following form. On the one hand, the delay cost $\mathcal{L}_{delay}(t)$ incurred in making a decision urges to make an early decision. On the other hand, the cost of making a bad prediction $\mathcal{L}_{prediction}$ is assumed to decrease over time, as the description of the incoming time series becomes richer. This decision making problem is \textit{under uncertainty}, since the hypothesis $h$ is capable of estimating the distribution of the possible outcomes $P(y|{\mathbf x}_t)$, at any time. 

\medskip
In practice, ECTS approaches trigger decisions at $\hat{t}$, hopefully the closest as possible to the optimal time $t^*$, at least in terms of cost: $\mathcal{L}(h({\mathbf x}_{\hat{t}}), \hat{t}, y) - \mathcal{L}(h({\mathbf x}_{t^\star}), t^\star, y)$ must be small. 
Triggering such a decision is an \textit{online} optimization problem, since $\hat{t}$ must be chosen based on a partial description ${\mathbf x}_t$ of the incoming time series ${\mathbf x}_T$ (with $t \leq T$), and the reward function can be defined as:

%\vspace{-3mm}

\begin{equation}
r({\mathbf x}_t,t,h({\mathbf x}_t),y)  =  \left\{
    \begin{array}{ll}
        - \mathcal{L}(h({\mathbf x}_t), t, y) & \mbox{if } t = \hat{t} \mbox{ or } t=T\\
        0 & \mbox{otherwise}
    \end{array}
\right.
\end{equation}

\noindent
where the risk equals to $0$ when no decision is made, given that the decision is forced at $t=T$ resulting in an important risk due to the delay cost.

\bigskip
In the rest of this section, and for readability reasons, the deadline $T$ is still considered as finite and known, as in the ECTS problem. In Section \ref{revocable_decision}, another setting is studied where $T$ is indeterminate, i.e. where the successive measurements are observed as a data stream. 

\bigskip
\noindent
{\bf \textit{Early decisions to be located in time}} constitute a more challenging problem, which consists of both making a decision for each incoming time series, but also predicting a \textit{time period} associated with the decision. 
For example, maintenance operations on hydroelectric dam turbines can only be performed when the electricity demand is at a low enough level. There are therefore periods where maintenance is possible and periods where this is not desirable. The objective here is to determine as early as possible \textit{whether} and during \textit{which period} it will be possible to shut down the turbines, within the day (if $[1,T]$ corresponds to one day).
In this case, the ground truth $(y, (s,e))$ consists of a class $y \in \mathbb{Y}$, associated with a certain time period $[s,e]$, defined by a \textit{start} timestamp $s \in [1,T]$ and a \textit{end} timestamp $e \in [s,T]$. %, such that $s \leq e$. 
%\Antoine{In the hydroelectric dam turbines example, There are therefore periods where maintenance is possible and periods where this is not desirable. }

At testing time, the objective is twofold: triggering the decision as early as possible, while also predicting the associated time period $[s,e]$.
Let us consider a decision denoted by $(h({\mathbf x}_{\hat{t}}), (\hat{s}, \hat{e}))$,
where $h({\mathbf x}_{\hat{t}})$ is the class predicted at $\hat{t}$ (the triggering time), and $[\hat{s}, \hat{e}]$ is the associated predicted time period.
The loss function $\mathcal{L}$ has to be redefined as a function of the following parameters:

%\vspace{-3mm}

\begin{equation}
    \mathcal{L} \left ( \underbrace{(h({\mathbf x}_{\hat{t}}), (\hat{s}, \hat{e}))}_{predictions}, \underbrace{\hat{t}}_{triggering\:time}, \underbrace{(y, {{(s,e)}})}_{ground\:truth} \right )
\label{eq:loss-function-local-decision}     
\end{equation}

\begin{figure}[htbp!]
\centering
\includegraphics[width=0.7\linewidth]{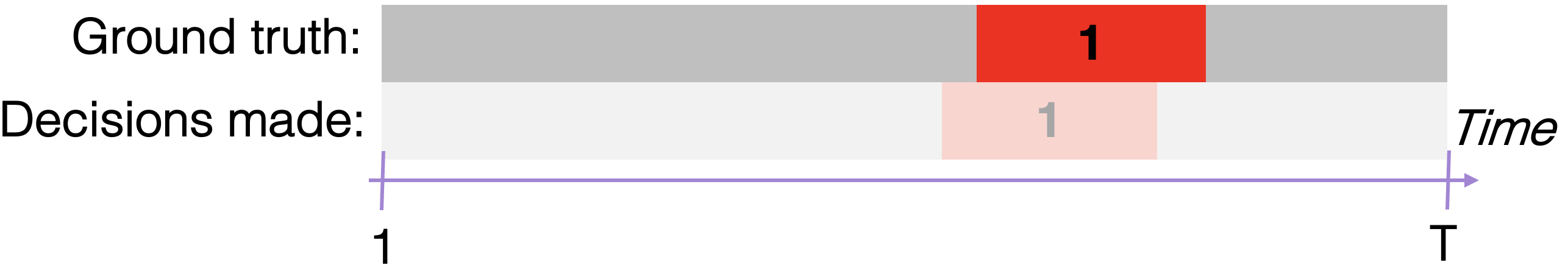}
%\vspace{-4mm}
\caption{Example of a time-lagged decision.}
\label{loss-local-decision}
\end{figure}

\medskip
The loss function $\mathcal{L}$ needs to be specified further, depending on the considered application. 
In general, this loss function should account for two aspects: (\textit{i}) the \textit{quality} of the predictions ; (\textit{ii}) the \textit{time overlap} between the decisions made and the true decisions. 
For example, Figure \ref{loss-local-decision} shows a situation where the decision made is correct, since the predicted class (see the second line) matches the ground truth (see the first line). 
But these two decisions do not coincide exactly in time, as the predicted time period is earlier than the ground truth. 

%\vspace{-3mm}

%\textcolor{gray}{Other situations are depicted below, which illustrate the decision costs involved (see Figures \ref{mapping-rule-max-delay} to \ref{loss-missing-delete}).}

\medskip
\noindent
By extension, ML-EDM considers {\bf \textit{multiple early decisions to be located in time}} (i.e. in the time period $[1,T]$).
, which is necessary in numerous applications.
For example, consider a set of servers used to trade on a stock exchange platform (where $[1,T]$ corresponds to the platform's hours of operation during the day).
For each server, key performance indices (e.g., CPU, RAM, network) are recorded over time.
The ground truth consists of a sequence of states (e.g., overload or nominal) associated with the corresponding time periods. 
In this application, the task is to detect overload periods as early as possible. 
%As an another illustration, in finance, one may be interested in predicting the time periods of the ups and of the downs of some stocks, given a time limit.

\medskip
Thus, in this problem, the true decisions $\{y_i, {{(s_i,e_i)}}\}_{i=1}^{k_{\mathbf x}}$ consists of a sequence of varying length $k_{\mathbf x}$, which is specific to each individual ${\mathbf x}$.  
Each element of this sequence is a decision to be \textit{located in time}, which consists of a class $y_i \in \mathbb{Y}$ associated with a certain time period $[s_i,e_i]$. 
For a given individual ${\mathbf x}$, the time periods $\{{{(s_i,e_i)}}\}_{i=1}^{k_{\mathbf x}}$ constitute a \textit{time partition}, each interval $[s_i,e_i]$ being associated with the true class $y_i$ (e.g. in predictive monitoring, this time partition would correspond to the successive states, up or down, of a given device).

\medskip
%, 
Here, the online optimization problem to be addressed is more complex than the previous one, since it consists in triggering a sequence of decisions as soon as possible, without knowing the number of true decisions $k_{\mathbf x}$, and also ignoring the time periods associated with each true decision. 
Let us consider that 
a ML-EDM approach triggers a sequence of decisions $\{h({\mathbf x}_{\hat{t}_{i'}}), (\hat{s}_{i'}, \hat{e}_{i'})\}_{i'=1}^{\hat{k}_{\mathbf x}}$ ; 
where $\hat{k}_{\mathbf x}$ is the number of decisions made ; 
%where two successive predictions must be different, such that $h({\mathbf x}_{\hat{t}_{i'}}) \neq h({\mathbf x}_{\hat{t}_{i+1'}}), \: \forall i' \in [1, \hat{k}_{\mathbf x}-1]$ ;   
where $\{\hat{t}_{i'}\}_{i'=1}^{\hat{k}_{\mathbf x}}$ represents the associated triggering times ; 
and where $\{(\hat{s}_{i'}, \hat{e}_{i'})\}_{i'=1}^{\hat{k}_{\mathbf x}}$ represents the predicted time periods associated to the decisions which forms a partition of the time period $[1,T]$. 
In the scenario of \textit{multiple early decisions to be located in time}, a loss function $\LL$ needs to be defined as a function of the following parameters:

\begin{equation}
    \LL \left ( \underbrace{\{h({\mathbf x}_{\hat{t}_{i'}}), (\hat{s}_{i'}, \hat{e}_{i'})\}_{i'=1}^{\hat{k}_{\mathbf x}}}_{predictions}, \underbrace{\{\hat{t}_{i'}\}_{i'=1}^{\hat{k}_{\mathbf x}}}_{triggering\:times}, \underbrace{\{y_i, {{(s_i,e_i)}}\}_{i=1}^{k_{\mathbf x}}}_{ground\:truth} \right )
\label{eq:double-loss-function}     
\end{equation}

This equation shows the loss function used to evaluate an approach after the deadline $T$, when predictions have been taken for all instants in the time period $[1,T]$.  
The loss function $\LL$ can be expressed in many different ways, depending on the application considered. 
In practice, \textit{mapping rules} need to be defined to match the decisions made to the true ones (see appendix in Section \ref{appendix1}).

% partie sur le revocable ou non ... 

Note that the problem of making predictions for all instants in $[1, T]$ points to the issue as whether decisions made can be revoked, or not, before $T$. In case decisions are irrevocable, once a decision has been made, let us say $(y, (s, e))$, then it is no longer possible to change the prediction of the class for all times $ t \in (s, e)$. This renders the optimization problem dependent upon previous decisions, and it becomes more constraining for application cases. Revocable decision are studied in Section \ref{revocable_decision}.

%{In fact, the problem is even more involved. Indeed, it points to the issue as whether decisions are revocable or not. In case decisions are irrevocable, then once a decision has been made, let us say $(y, (s, e))$, then it is no longer possible to change the prediction of the class for all times $ t \in (s, e)$. This renders the decision making problem dependent upon previous decisions, and the optimization problem becomes more complicated.}

\RemoveForShortVersion{
\bigskip
One of the major issues of ML-EDM is to make the time periods associated with the decisions taken $\{(\hat{s}_{i'}, \hat{e}_{i'})\}_{i'=1}^{\hat{k}_{\mathbf x}}$ coincide with those associated with the true decisions $\{{{(s_i,e_i)}}\}_{i=1}^{k_{\mathbf x}}$. In this respect, two situations can be distinguished which have an important impact on the triggering times $\{\hat{t}_{i'}\}_{i'=1}^{\hat{k}_{\mathbf x}}$:
\begin{itemize}
    \item In the case of a decision making problem \textit{under ignorance}, the occurrence of the next true decision is independent of data observed before it occurs.  
    For example, in a predictive monitoring application, this means that there is no premise in data correlated to future failures.
    In this case, the triggering times $\hat{t}_{i'}$ will all happen after $s_i$, that is the beginning of the time period associated with the corresponding true decision (a situation where $\hat{t}_{i'} > s_i$ is called a \textit{negative} prediction horizon).
    \item In the case of a decision making problem \textit{under uncertainty}, data observed at the current time ${\mathbf x}_t$ may provide useful information to predict the occurrence of the next true decision. The partition of true decisions $\{{{(s_i,e_i)}}\}_{i=1}^{k_{\mathbf x}}$ is generated by a stochastic process which can be modeled conditionally to the observed data (as in \cite{frazier2007sequential} which deals with a stochastic decision deadline). 
    In a predictive monitoring application, this means that observed data include premises which are correlated to future failures. The triggering moments $\hat{t}_{i'}$ can then be located before $s_i$ (a situation where $\hat{t}_{i'} < s_i$ is called a \textit{positive} prediction horizon).
\end{itemize}
\noindent
ML-EDM aims at designing approaches which correctly manage both situations, i.e. decision making under \textit{ignorance} or \textit{uncertainty}, without being informed about it. If the observed data contains useful information for predicting the timing of the next true decision, the approach should be able to exploit it. If not, the approach should still perform properly by triggering decisions under negative prediction horizons.
}

\bigskip
\noindent
The {\bf \textit{deadline $T$}} after which decisions are forced is an important component, that takes different forms depending on the problem. 
In the simple case of ECTS, only one decision needs to be made before the incoming time series is complete. Thus, the deadline $T$ is defined as the \textit{maximum size} of the input series, which is known in advance during training.  
By contrast, in the more complex case of ML-EDM where multiple decisions to be located in time must be taken, the deadline $T$ is defined as a \textit{maximum delay} allowed to detect the start of a true decision (i.e. a bound on negative decision horizons). In practice, two situations can be distinguished:

\begin{itemize}

    \item Some applications do \textit{not support the absence of decision},  and the entire considered time period must be partitioned by the successive decisions. 
    This is the case for instance when moderating content on social networks, where discussions are continuously going on between users and where each part of these discussions must be classified as \textit{appropriate} or \textit{not} (see Section \ref{sec_usecase_social_network}). 
    In this case, no decision is not allowed, and all decisions are subject to the cost $\mathcal{L}_{delay}$ and thus constrained by the deadline $T$.

    \item By contrast, in some applications, a \textit{nominal operating state} exists which is almost permanent, and for which there is no decision deadline. This is for instance the case in predictive maintenance applications, where there is no urgency or even a deadline to detect the absence of failure. In this case,  the delay cost $\mathcal{L}_{delay}$ and also the deadline $T$ apply only to the other decisions (e.g. failures categorized by severity level) excluding the nominal state.  
\end{itemize}

At the end, ML-EDM aims to develop approaches which allow for easy adaptation to all cases, whether the deadline $T$ is applicable to all decisions, or whether there exists a nominal operation state which bypasses this deadline.

% --------------------------------------------------------------------------

\bigskip
\noindent
\textbf{Question B - }\textit{How to learn a triggering strategy from data?}\\
~\\
As a summary, this section shows that learning a triggering strategy follows the usual general principles of Machine Learning approach, with the particularity to consider \textit{time-sensitive} loss functions (i.e. which depend on when decisions are triggered, as in Equations \ref{eq:loss-function-ECTS}, \ref{eq:loss-function-local-decision} and \ref{eq:double-loss-function}). 

\bigskip
In practice, the optimal triggering strategy is not available and it must be approximated by a learned function, such as $Trigger^\gamma \approx  Trigger^*$, where $\gamma \in {\Gamma}$ is a set of parameters to be optimized within the space of parameters ${\Gamma}$ of a chosen family of triggering strategies.    

In addition, the hypothesis $h$ is supposed to be learned previously during the training phase, making the system capable of predicting $y$ at any time $t \in [1,T]$. This hypothesis is defined by a set of parameters $\theta \in \Theta$. 

To illustrate what a triggering strategy is, let us consider an example from the ECTS literature.  
The \textsc{SR} approach, described in \citep{mori2017early}, involves 3 parameters $(\gamma_1, \gamma_2, \gamma_3)$ to decide if the current prediction $h({\mathbf x}_t)$ must be chosen (output $1$) or if it is preferable to wait for more data (output $0$):

%\small
\begin{equation}
Trigger^\gamma \left ( h({\mathbf x}_t) \right )  =  \left\{
    \begin{array}{ll}
        0 & \mbox{if } \gamma_1 p_1 + \gamma_2 p_2 + \gamma_3 \frac{t}{T} \le 0 \\
        1 & \mbox{otherwise}
    \end{array}
\right.
\label{eq:mori_trigger_function}
\end{equation}
where $p_1$ is the largest posterior probability estimated by $h$, $p_2$ is the difference between the two largest posterior probabilities, and the last term $\frac{t}{T}$ represents the proportion of the incoming time series that is visible at time $t$.  
The parameters $\gamma_1, \gamma_2, \gamma_3$ are real values in $[-1, 1]$ to be optimized, as described more generally in the following. 

\bigskip
In the simple case of ECTS, a single decision has to be made for each time series ${\mathbf x \in \mathbb{X}}$ (see Equation \ref{eq:loss-function-ECTS}). Thus, the \textit{risk} associated with any triggering strategy $Trigger^\gamma$ belonging to any family $\Gamma$, is defined as follows, given the previously learned hypothesis $h^\theta$ within the family $\Theta$: 

%\begin{equation}
%\begin{split}
%  R&(Trigger^\gamma| h^\theta ) = \expectancy{\mathbb{X}, \mathbb{Y} }{\mathcal{L}(h^\theta({\mathbf x}_{\hat{t}}), \hat{t}, y)}{} \\
%  &= \int_{\mathbf{x} \in \mathbb{X}}\int_{y \in \mathbb{Y}}\mathcal{L}\left (h^\theta({\mathbf x}_{\hat{t}}), \hat{t}, y \right) \, \text{d}P(\mathbf{x},y) 
%\end{split}
%\label{eq:risk_ECTS}
%\end{equation}
\begin{equation}
\begin{split}
  R&(Trigger^\gamma| h^\theta ) = \expectancy{\mathbb{X}, \mathbb{Y} }{\mathcal{L}(h^\theta({\mathbf x}_{\hat{t}}), \hat{t}, y)}{}
\end{split}
\label{eq:risk_ECTS}
\end{equation}
where $\hat{t}$ is determined by $\gamma$, the parameters of the triggering strategy. 

\smallskip
Similarly, the risk can be defined in the more complex case of {\bf \textit{multiple early decisions to be located in time}}.
Let $T_{part}$ be the set of all possible partitions of the time domain $[1,T]$, having a varying number of time intervals $k$.
The risk can be defined as:

%\vspace{-2mm}

%\begin{equation}
%\begin{split}
%  &R(Trigger^\gamma|h^\theta) = \\
%  &\expectancy{\mathbb{X}, T_{part}, \mathbb{Y}^k }{\LL(\{h^\theta({\mathbf x}_{\hat{t}_{i'}}), (\hat{s}_{i'}, \hat{e}_{i'})\}, {\{\hat{t}_{i'}\}}, \{y_i, {{(s_i,e_i)}}\})}{} \\
%  &= \hspace{-2mm} \int \hspace{-2.5mm} \int \hspace{-2.5mm} \int_{\substack{ \mathbf{x} \in \mathbb{X} \\ \{(s_i,e_i)\} \in T_{part} \\ \{y_i\} \in \mathbb{Y}^{k} }} \hspace{-2.5mm} \LL \left (\{h^\theta({\mathbf x}_{\hat{t}_{i'}}), (\hat{s}_{i'}, \hat{e}_{i'})\}, {\{\hat{t}_{i'}\}}, \{y_i, {{(s_i,e_i)}}\}\right) \\
%  &\hspace{55mm}dP(\mathbf{x},\{(s_i, e_i)\},\{y_i\})
%\end{split}
%\label{eq:double-loss}
%\end{equation}
%
\begin{equation}
\begin{split}
  &R(Trigger^\gamma|h^\theta) = \\
  &\expectancy{\mathbb{X}, T_{part}, \mathbb{Y}^k }{\LL(\{h^\theta({\mathbf x}_{\hat{t}_{i'}}), (\hat{s}_{i'}, \hat{e}_{i'})\}, {\{\hat{t}_{i'}\}}, \{y_i, {{(s_i,e_i)}}\})}{}
\end{split}
\label{eq:double-loss}
\end{equation}
where ${\{\hat{t}_{i'}\}}$ and $\{(\hat{s}_{i'}, \hat{e}_{i'})\}$ are determined by $\gamma$, and given $h^\theta$.
%
%\smallskip
In Equation \ref{eq:double-loss}, the risk is an expectancy on three random variables, drawing triplets from the join distribution $P(\mathbf{x},\{(s_i, e_i)\},\{y_i\})$.   
The first element corresponds to the input data\footnote{Notice that the notation $\mathbf{x} \in \mathbb{X}$ in Equations \ref{eq:risk_ECTS} and \ref{eq:double-loss} is an abuse that we use use to simplify our purpose. In all mathematical rigor, the measurements observed successively constitute a family of time-indexed random variables $\mathbf{x} = (\mathbf{x}_t)_{t \in [1,T]}$. This stochastic process $\mathbf{x}$ is not generated as commonly by a distribution, but by a filtration $\mathbb{F} = (\mathcal{F}_t)_{t \in [1,T]}$ which is defined as a collection of nested $\sigma$-algebras \cite{klenke2013} allowing to consider time dependencies. Therefore, the distribution $P(\mathbf{x},\{(s_i, e_i)\},\{y_i\})$ should also be re-written as a filtration.}, which is an individual $\mathbf{x} \in \mathbb{X}$. 
The two other consist of the ground truth, which is composed of: (\textit{i}) a partition of the time domain $\{(s_i,e_i)\} \in T_{part}$ with a particular number of time intervals, denoted by $k \in [1,T]$ ; (\textit{ii}) and a set of class labels $\{y_i\} \in \mathbb{Y}^{k}$ for each time interval. 

\smallskip
Now, the objective is to approximate the optimal triggering strategy $Trigger^*$ by finding $\gamma^* \in \Gamma$ which minimizes the risk, such that: 
\begin{equation}
    \gamma^* = \argmin_{\gamma \in \Gamma} R(Trigger^\gamma | h^{\theta})
\end{equation}

%\vspace{-2mm}

\noindent
% https://en.wikipedia.org/wiki/Filtration_(probability_theory)
% https://en.wikipedia.org/wiki/Probability_space 
The joint distribution $P(\mathbf{x},\{(s_i, e_i)\},\{y_i\})$ is unknown, thus Equation \ref{eq:double-loss} can not be calculated ; however a training set $\mathcal{S}$ which samples this distribution is supposed to be available.   
The risk can be approximated by the \textit{empirical risk} calculated on the training set $\mathcal{S} = \{{\mathbf x}^j, \{y^j_i, {{(s_i^j,e_i^j)}}\} \}_{j \in [1,n], i \in k_{\mathbf{ x}_j}}$, as follows:
\begin{equation}
\begin{split}
  &R_{emp}(Trigger^\gamma |  h^{\theta}) =\\
  &\frac{1}{n} \sum^n_{j=1} \LL \left (\{h({\mathbf x}^j_{\hat{t}_{i'}}), (\hat{s}^j_{i'}, \hat{e}^j_{i'})\}, {\{\hat{t}_{i'j}\}}, \{y_i^j, {{(s_i^j,e_i^j)}}\} \right )
\end{split}
\label{risk_emp_8}
\end{equation}

\noindent
where $\hat{t}_{i'j}$ is the triggering time of the \textit{i-th} made decision of the \textit{j-th} individual.

\bigskip
At the end, training a ML-EDM approach can be viewed as a \textit{two-step} Machine Learning problem: (\textit{i}) \textit{first}, the hypothesis $h^{\theta}$ must be learned in order to predict the most appropriate decision $h^{\theta}(\mathbf{x}_t)$, at any time $t \in [1,T]$ ; (\textit{ii}) \textit{second}, the best triggering strategy defined by $\gamma^*$ must be learned, given the hypothesis $h^{\theta}$ and given the family $\Gamma$, such that: 
\begin{equation}
    \gamma^*  = \argmin_{\gamma \in \Gamma} R_{emp}(Trigger^\gamma| h^{\theta})
\label{eq_learn_optimal_strategy}    
\end{equation}

% ---------------------------------------------------------------------------------

\bigskip
\noindent
\textbf{Question C - }\textit{Can a triggering strategy be learned by Reinforcement Learning?}\\

\noindent
To sum up, this section shows that 
learning a triggering strategy of a ECTS approach  
can be cast as a Reinforcement Learning (RL) problem, with rewards well chosen, and it might be expected that provided with sufficient training, RL learning may end up with a good approximation of an efficient decision function. 

%However, using this learning strategy would \Alexis{(would a nuancer?)} be exceedingly costly in terms of the required quantity of training examples as compared to providing at once a well-thought decision criterion. 

\bigskip
Reinforcement learning \cite{sutton2018reinforcement} aims at learning a function, called a policy $\pi$, from states to actions: $\pi: {\cal S} \rightarrow {\cal A}$. Rewards can be associated with transitions from states $s_t \in {\cal S}$ to states $s_{t+1} \in {\cal S}$ under an action $a \in {\cal A}$. Rewards are classically denoted $r(s_t, a, s_{t+1}) \in \mathbb{R}$. 
In all generality, the result of an action $a$ in state $s_t$ may be non deterministic and one among a set (or space) of states. The optimal policy $\pi^\star$ is the one that maximizes the expected gain from any state $s_t \in {\cal S}$. This gain, denoted $R_t$ starting from the sate $s_t$, is defined as a function of the rewards from that state (e.g. a discounted sum of the rewards received).   
In order to learn a policy, value functions can be considered, such as the state-value function $v_\pi(s)$ classically defined as:
\begin{equation}
\begin{split}
    v_\pi(s_t) \; &\doteq \; \mathbb{E}_\pi[R_t \, | \, s_t] = \; \sum_{a \in {\cal A}} \pi(a | s_t) \, \sum_{s_{t+1}, r} \\ 
    &p(s_{t+1},r \, | \, s_t,a) \, \bigl[r(s_t,a,s_{t+1}) \, + \, \gamma \, v_\pi(s_{t+1}) \bigr] 
\end{split}
\label{eq_rl_v_function}
\end{equation}
where $\mathbb{E}_\pi[\cdot]$ denotes the expected value of a random variable given that the agent follows the policy $\pi$ and $t$ is any time step. In the case of a non deterministic policy, $\pi(a | s_t)$ denotes the probability of choosing action $a$ in state $s_t$ and $p(s_{t+1},r \, | \, s_t,a)$ the probability of reaching state $s_{t+1}$ and receiving the reward $r$ given that the action $a$ has been chosen in state $s_t$. And $\gamma$ is a discounting factor: $\gamma < 1$.  

In our case, the agent aims to learn a triggering strategy given the previously learned classifier $h^\theta$, and the state $s_t = (t, \mathbf{x}_t)$ is the current time $t$ and the observed data at current time. The instantaneous reward $r(s_t,a)$ only depends on the current state $s_t$ and the action taken $a$ (i.e. prediction now, or postponed to a later time).   
Finally, the discounted factor $\gamma$, usually present in RL for reasons of convergence over infinite episodes, is equal to 1 in our case, since we always deal with finite episodes with forced decisions after a maximum delay. So that the equation (\ref{eq_rl_v_function}) simplifies to: 
%\vspace{-1.5mm}
\begin{equation*}
    v_\pi(s_t) \; = \; \sum_{a \in {\cal A}} \pi(a | s_t) \, r(s_t,a) \, + \, v_\pi(s_{t+1}) 
\end{equation*}
when, during learning, the agent takes a decision, it updates the value of the state $s_t$ using:
\begin{equation*}
    v_\pi(s_t) \; = \; r(s_t,a) \, + \, v_\pi(s_{t+1}) 
\end{equation*}
where $s_{t+1}$ is the state after having taken the action $a$ in state $s_t$. 

As the equation above shows, the core observation in RL is that the value function for a state $s_t$ (i.e. an estimation of the expected gain from that state) is related to the value function of states $s_{t+1}$ that may be reached from $s_t$. In that way, information gathered further down a followed path can be back-propagated to previous states thus allowing increasingly better decisions from those states to be made. 

\begin{figure}[htbp!]
\centering
 \includegraphics[width=0.75\linewidth]{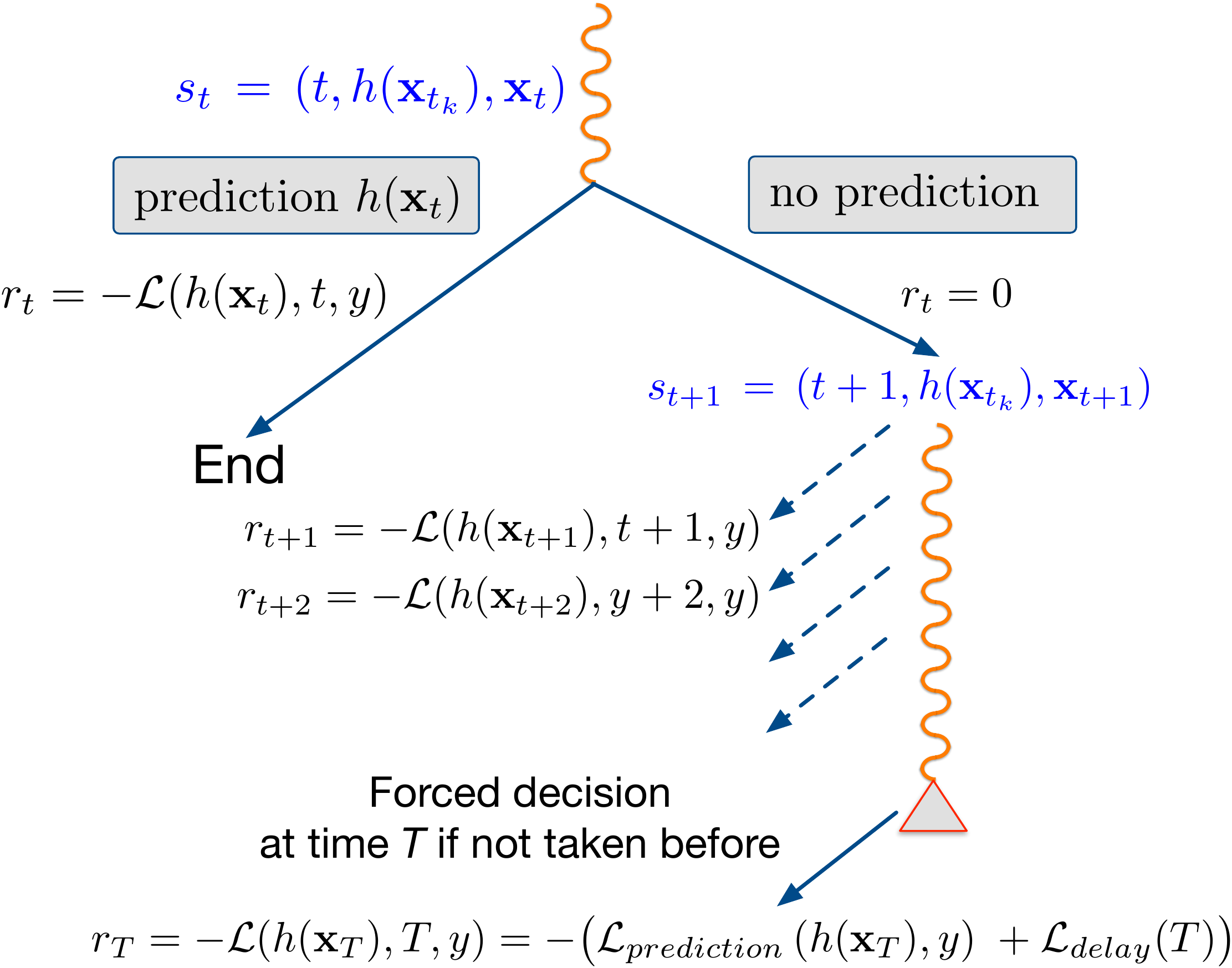}
\caption{A part of a ECTS ``game'' when learning an optimal policy while ``playing'' a training time series. When a prediction is made, the game stops, otherwise it continues until a prediction is made or the term of the episode is reached.}
\label{fig_RF_ECTS}
\end{figure}
For instance, in game playing, rewards may happen both during play (e.g. the player just lost a pawn) and at the end of the game (e.g. the player is chess mate).
Similarly, one could cast the ECTS problem as a RL problem where, at each time step, the ``player'' is in state $s_t = (t, h(\mathbf{x}_{t_k}), \mathbf{x}_t)$  and should choose between making a prediction (e.g. $h(\mathbf{x}_t)$) with an associated reward: $r_t = - \mathcal{L}(h(\mathbf{x}_{t}), {t}, y) = -\mathcal{L}_{prediction} \left( h(\mathbf{x}_t),y \right) \; -  \mathcal{L}_{delay}(t) $ or postpone the  decision, with no immediate associated reward, that is $r_t=0$. If no decision has been made before the term of the episode (e.g. when $t = T$) a decision is forced (see Figure \ref{fig_RF_ECTS}).
Provided with enough time series to train on, and sufficient training in the form of ``playing'' these time series, a reinforcement learning agent may end up with a policy $\widehat{\pi}$ that approximates a good early triggering strategy, one that would converge over time, after a very large number of ``plays'' on the training time series, to the optimal decision function $\pi^\star$ (See Equations \ref{eq:trigger_function} and \ref{eq_learn_optimal_strategy}
). 

\medskip
The RL framework is very general. It uses immediate and delayed rewards. As shown in this section, there is in principle no obstacle to apply RL to the learning of a good triggering strategy. However, if used directly, the generality of RL is paid for by a need for a large number of ``experiments''. In addition, the state space is continuous in the case of the ECTS problem, thus an interpolating functions must be used in order to represent the values such as $v_\pi(s)$ and this entails the choice of a family of functions and setting their associated parameters. 
Another approach, the one favored in the current literature for ECTS \cite{achenchabe2021MLj}, is to choose functions for representing the expected values of decision times, and thus providing a ground for the triggering strategy. 

%This has the merit of integrating prior knowledge of the trade-off between earliness and precision, but the disadvantage of having to choose a priori a method of estimating this trade-off which may be biased.

This has the merit of incorporating prior knowledge of the trade-off between earliness and accuracy, at the cost of making modelling choices that may bias the method of estimating the expected future cost.

The respective performances, merits and limits of both approaches should be studied empirically by a comparison of RL based ECTS approaches, such as \cite{martinez2020}, with approaches that explicitly exploit the form of the optimization criterion designed for ECTS as in \cite{achenchabe2021MLj}.

\section{Origin of the delay cost}
\label{decision_costs}
%-------------------------------------------------------------------
%-------------------------------------------------------------------

ML-EDM approaches aim to trigger decisions at the right time, by reaching a good trade-off between the \textit{earliness} and the \textit{accuracy} of their decisions. 
To achieve this, a balance must be found between penalizing {late decisions} and penalizing {prediction errors}. 
\textit{Decision costs} are key to make this antagonistic trade-off choice, as they allow us to evaluate the cost of waiting for new measures vs. the cost of making a decision now. 
In Section \ref{definition-ML-EDM}, decision costs are involved starting from Equation \ref{eq:trigger_function} in the loss function $\mathcal{L}$ and they have an important impact on the entire path of the description of the ML-EDM problem. The objective of this section is to understand the deep origin of the delay cost. 

%-------------------------------------------------------------------
%\bigskip
%\noindent
%\textit{The delay cost originates from the task parallelization}

\bigskip
\noindent
The \textit{delay cost} represents the cost of postponing a decision (see the function $\mathcal{L}_{delay}$ in Equation \ref{eq:loss-function-ECTS}).
In the particular case of ECTS problems, the delay cost is present in all the works described in scientific literature. But it can be explicitly defined as in \cite{mori2015early, achenchabe2021MLj}, or implicitly as in most approaches. 
For instance, the authors in \cite{xing2009early} trigger all the decisions at the \textit{minimum prediction length}, which correspond to the early moment such that no prediction differs from those applied to the full-length training time series (based on a KNN classifier). 
This approach thus \textit{implicitly} assumes that the delay cost is very low, by favoring the accuracy of decisions at the expense of their earliness. 
In \cite{mori2019early}, the authors propose to model the trade-off between earliness and accuracy as a multi-objective criterion and explore the Pareto front of multiple dominant solutions. 
This approach is useful in applications where earliness and accuracy can not be evaluated in a commensurable way, and it provides a collection of optimal solutions each corresponding to a particular value of the delay cost. 

\begin{figure}[htbp!]
\centering
\includegraphics[width=0.7\linewidth]{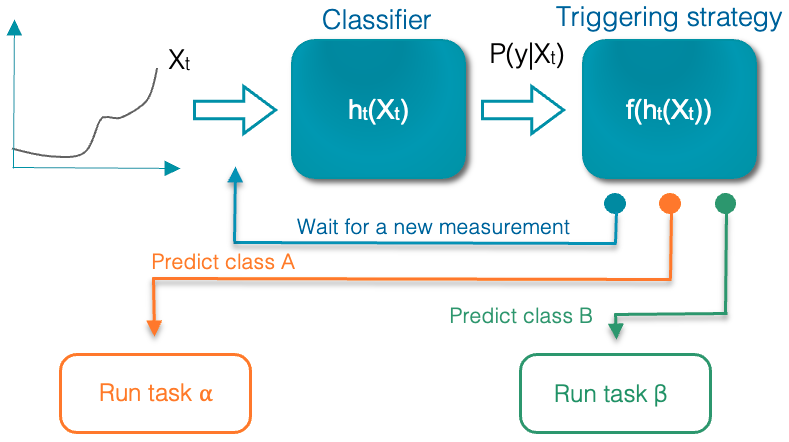}
\caption{Tasks to be performed after the triggering of a decision.}
\label{fig:cost-origin}
\end{figure}

\bigskip
For a better understanding, let us examine what happens once a decision is triggered in the simple ECTS problem.  
Figure \ref{fig:cost-origin} represents a \textit{classifier} and a \textit{triggering strategy}. 
At each time step $t \in [0,T]$, the classifier predicts the conditional distribution $P(y|{\mathbf x}_t)$ based on the input incomplete time series ${\mathbf x}_t = \langle x_0, x_1, \ldots, x_t \rangle$. 
Then, the triggering strategy  either decides to \textit{postpone} the decision until a new measurement $x_{t+1}$ is available, or to \textit{trigger} the decision by predicting the class value.   
In this first scenario, let us consider that triggering a decision at time $t$ implies performing a given \textit{task} (namely $\alpha$ or $\beta$) which depends on the predicted class (respectively $A$ or $B$).   

Given that this task ($\alpha$ or $\beta$) must be completed \textit{before} the deadline $T$, the problem is to determine how the cost of performing this task evolves depending on the trigger time $t$. 
In practice, the delay cost $\mathcal{L}_{delay}$ takes the form of a parametric function (e.g., a constant \cite{xing2009early}, linear \cite{achenchabe2021MLj} or exponential \cite{beibel2000} function), 
whose form characterizes the additional cost to delay the execution of the tasks.

\medskip
A \textit{constant cost}, one where there is no penalty associated with delaying the decision, would mean that these tasks are achievable in an arbitrarily short time $T-t < \epsilon$. 
In practice, an irreducible amount of time is needed to perform the tasks using a single worker. To reduce this time, the tasks need to be parallelized using several workers, incurring an extra-cost when building the global result from sub-tasks. 
Formally, a constant delay cost would mean that the tasks are \textit{infinitely parallelizable}, i.e. they can be divided into \textit{independent} and \textit{arbitrarily small} sub-tasks, and that there is no extra-cost in building the global result.

%\medskip
%In practice, parallelizing a task necessarily induces a cost which increases with the number of workers.  
%Let's take the example of the handwritten copy of a book, as it was done by scribe monks before the invention of printing.  
%To complete this task as quickly as possible, it is wise to have several scribes each copying a packet of pages and then assembling these packets to produce the final copy of the book. 
%If now the time to complete this task is infinitely short, the same strategy consists in mobilizing an infinite number of scribes, each one being in charge of writing an infinitesimal part of the text (i.e. a dot on a blank sheet). The rebuilding of the copy from the work performed by the scribes would be infinitely costly, since it consists of an incredibly hard and very meticulous scrapbooking work. 

%\medskip
%This (absurd) example shows that parallelizing a task necessarily induces a cost which increases with the number of workers.  
%Therefore, 

%The cost of performing the tasks increases monotonically with the triggering time $t$, hence the diminishing remaining time, and it may depend on the decision made (i.e. the predicted label). 

\medskip
More generally in ML-EDM problems, the delay cost $\mathcal{L}_{delay}$ is necessarily an increasing function (monotonic or piecewise) depending on the time remaining before the decision deadline, and it may depend on the decision made (i.e. the predicted label). In addition, it should tend to $+ \infty$ when the time remaining to perform these tasks $T-t$ tends to zero \cite{beibel2000}. 
For example, this delay cost may be modeled by $\mathcal{L}_{delay}(t) =  1/(T-t)^\alpha$, with a single parameter $\alpha$ which influences the increase in cost when $(T-t) \rightarrow 0$.    

%-------------------------------------------------------------------
%-------------------------------------------------------------------
\section{Learning tasks}
\label{learning_tasks}
%-------------------------------------------------------------------
%-------------------------------------------------------------------

The formal definition of ML-EDM provided in Section \ref{definition-ML-EDM} involves the ground truth. However, in many applications, it is extremely hard or costly to obtain, especially in the case of anomaly detection (e.g. fraud, cyber-attacks, predictive maintenance). In these application domains, there are several issues: (\textit{i}) labels can be extremely expensive to obtain as they each require an examination from an expert ; (\textit{ii}) the labels provided by experts can be uncertain ;  and (\textit{iii}) the class of anomalous observations is often poorly represented and drifts over time. For example, cyber-attack techniques are very diverse and change with time.  
Faced with these difficulties, anomaly detection is often addressed using unsupervised approaches, by assuming that the anomalies are outliers. %
In this case, the problem comes down to modeling the normal behavior of the system, if possible using historical data that are cleaned of anomalies. 
Then, it is necessary to define the notion of outlier to be able to assign an eccentricity score to the new observations.
Note that this type of modeling can be considered as a \textit{first step} to manage non-stationarity, since in this case the stationarity assumption only concerns the normal behavior of the system (this assumption could be removed in future work).

%-------------------------------------------------------------------
\bigskip
\noindent
{\bf Challenge \#1: \\extending non-myopia to unsupervised approaches}~\\
An unsupervised ML-EDM problem could be to decide, as soon as possible, whether a partially observed time series $\langle x_1, x_2, \ldots, x_t \rangle$ will be an outlier (or not) when fully observed at time $T$. 
In this case, the \textit{accuracy vs. earliness} trade-off still exists. 
On the one hand, an early detection is inaccurate by nature because the outlier series is unreliably detected, based on few observed measurements. 
On the other hand, delaying the detection of anomalies can be very costly.
For instance, a cyber-attack which is not detected immediately gives time to the hakers to exploit the security hole found.  
Designing ML-EDM approaches to tackle  \textit{unsupervised} learning tasks is challenging in several respects: (\textit{i}) learning a triggering strategy with the goal of achieving a good trade-off between earliness and accuracy of its decisions cannot be achieved in the Machine Learning framework as described in the section \ref{definition-ML-EDM} and should be formalized in another way ; (\textit{ii}) developing unsupervised \textit{non-myopic} approaches is very difficult, as the training set does not contain anomalous series, thus the triggering strategy  cannot learn from their continuations.

\bigskip
The extension of ML-EDM both to \textit{online} scenarios (see Section \ref{stream}) and to \textit{unsupervised} tasks is of particular interest, because combined they would enable a new generation of \textit{monitoring systems} \cite{abellan2010review} to be developed.  In this case, the learning task would consist in detecting online the start and end of the outlier chunks: (\textit{i}) without requiring labels to learn the model ;
(\textit{ii}) by considering the trade-off between \textit{accuracy} and \textit{earliness} to trigger the decisions at the right time. 

%-------------------------------------------------------------------

\bigskip
\noindent
{\bf Challenge \#2: addressing other supervised learning tasks}~\\
The formal description of ML-EDM proposed in Section \ref{definition-ML-EDM} is generic, in the sense that the type of the target variable $y$ can easily be changed. 
By definition, the ECTS approaches in the literature are limited to \textit{classification} problems, but they could naturally be extended to other supervised learning tasks. 
For instance, predicting a \textit{numerical} target variable from a time series is a problem known as \textit{Time Series Extrinsic Regression} (TSER)\cite{tan2021time}. 
In some domains, TSER approaches are very useful and allow applications such as the prediction of the daily energy consumption of a house, based on the last week's consumption, temperature and humidity measurements. 
\textit{Early} TSER would consist of predicting the value of the numerical target variable as soon as possible, while ensuring proper reliability. 
Another example of a supervised task for which ML-EDM approaches could be developed is time series \textit{forecasting} \cite{chatfield2000time}. 
Basically, a forecasting model aims to predict the next measurements of a time series up to an horizon $\nu$,  $Y = \langle x_{t+1}, x_{t+2}, \ldots, x_{t+\nu} \rangle$ from the recent past measurements $X = \langle x_{t-w}, \ldots, x_{t-1}, x_{t} \rangle$.    
Using a forecasting model, in a an online and \textit{early} way, would consist of adapting the forecast horizon $t+\nu$ according to the observed values in $X$, by modeling the trade-off between the \textit{accuracy} and the \textit{earliness} of these predicted values.  

\bigskip
The ML-EDM problem described in Section \ref{definition-ML-EDM} should also be adapted to \textit{semi-supervised} learning, which is of great help when the ground truth is only partially available. 
More generally, the collected ground truth may be imperfect for various practical reasons, such as the labeling cost, the availability of experts, the difficulty of defining each label with certainty, etc.
This problem has recently gained attention in the literature through the field of \textit{Weakly Supervised Learning} (WSL) \cite{zhou2018brief} which aims to list these problems and provide solutions.
\RemoveForShortVersion{
As detailed in \cite{nodet2021weakly}, the collected labels may suffer from three main deficiencies: (\textit{i}) inaccuracy ; (\textit{ii}) non-adaptability ; (\textit{iii}) or even incompleteness. 
More precisely:
\begin{enumerate}
    \item[i)] \textit{Inaccuracy} of labels is commonly modeled as noise: the probability distribution that a label is corrupted may be uniform (i.e. completely at random), may depend on the class value (i.e. at random), or even it may depend on the instance by varying in the input space (i.e. not at random);
    \item[ii)] \textit{Non-adaptability} of labels gathers a variety of situations, such as Transfert Learning \cite{zhuang2020} and Multi-instance Learning \cite{Carbonneau_2018}, where the training labels may be available in another target domain or a sub-domain (i.e. proxy domain), the labels may come from a slightly different concept than the one to be learned (i.e. proxy labels), or labels can be associated with  individuals defined in a slightly different way (i.e proxy individual) ; 
    \item[iii)] \textit{Incompleteness} of labels is related to partially labelled training sets. The objective is to use the entire training set, including the unlabeled examples, to achieve better classification performance than learning a classifier only from labeled examples. Active Learning \cite{settles2009active}, Semi-Supervised Learning \cite{seeger2000learning}, Positive Unlabeled Learning \cite{bekker2020learning}, Self-Training \cite{ennaji2012self} and Co-training \cite{blum1998combining} are suitable techniques for this situation.  
\end{enumerate}
}

%-------------------------------------------------------------------
\bigskip
\noindent
{\bf Challenge \#3: early weakly-supervised learning}~\\ 
The extension of ML-EDM to weakly-supervised learning is an interesting challenge, as it would allow to better address applications where the ground truth has corruptions or is incomplete (which includes semi-supervised learning). 
However, the weakly-supervised learning is a very large domain with many types of supervision deficiencies to be studied. 
From a practical point of view, the priority is probably to extend ML-EDM to label noise, and more specifically to bi-quality learning \cite{nodet2020importance}, where the model is trained from two training sets: (\textit{i}) one trusted with few labels ; (\textit{ii}) the other, untrusted, with a large number of potentially corrupted labels. This would allow interesting applications, such as in cyber security where few labels are investigated by an expert, and the majority of labels are provided by rule-based systems. 
The major difficulty in designing \textit{bi-quality learning} ML-EDM approaches is to learn a triggering strategy  from these two training sets, which models the compromise between \textit{accuracy} and \textit{earliness} in a robust way to label noise. 
Another interesting avenue would be to adapt \textit{Active Learning} \cite{settles2009active} approaches to ML-EDM, with the goal of labeling examples which improve both \textit{accuracy} and \textit{earliness} of the decisions. Such approaches would be particularly helpful when early decisions have to be made, and when labeling examples is very costly as, again, it is the case in cyber security applications.

%-------------------------------------------------------------------
%-------------------------------------------------------------------
\section{Types of data}
\label{data_types}
%-------------------------------------------------------------------
%-------------------------------------------------------------------

The ML-EDM definition proposed in Section \ref{definition-ML-EDM} involves 
%data which describes a single individual through 
measurements (i.e. scalar values) acquired over time. 
%, denoted by ${\mathbf x}_t$. 
However, this is only for reasons of simplicity of exposition. Ideally,  ML-EDM approaches should be \textit{data type agnostic}, i.e. they should operate for any data type as long as measurements are made over time and decisions are online. 

Below, we outline data types that are present in applications where ML-EDM could be used.

%The considered type of data was not specified, because it is a generic type which must be adapted depending on the application. 
%Here are some examples of data types for which ML-EDM approaches could be applied: 

\begin{enumerate}
    
    \item[i)] \textit{Multivariate time series} consist of successive measurements each containing more than one numerical value. \RemoveForShortVersion{More formally, a multivariate time series of length $t$ is defined as ${\mathbf x}_t \, = \, \langle (x_1^1 \ldots x_1^k), \ldots, (x_t^1 \ldots x_t^k) \rangle$, where $k$ is the number of numerical values composing each time-indexed vector. For example, the data from an accelerometer can be represented by a multivariate time series composed of the three variables representing the accelerations along the X, Y, and Z axes.}  
    
    \item[ii)] More \textit{complex signals} exist, such as video streams which involve higher dimension. \RemoveForShortVersion{Typically, a video stream is composed of frames, i.e. the measurements are images. Each time-indexed frame can be considered as a signal that is indexed by the position $(x,y)$ of the pixels. The values of this multi-indexed signal are the pixel, defined by four values which encode the $RGB$ components and the luminance of each pixel.
    Each time-indexed frame can be considered as a 2D signal composed of 3 channels which describe the RGB components. In addition,  video streams could be multimedia when they include speech and transcript (text or sign language translation).}
    
    \item[iii)] \textit{Data streams} is another type of data which can contain both numeric and categorical variables \cite{bifet2018machine}. Successive measurements\RemoveForShortVersion{, also called \textit{tuples},} are received in an uncontrolled order and speed. \RemoveForShortVersion{For example, an Internet of Things (IOT) device such as a connected security camera may send measurements in an irregular flow. These measurements may be composed of categorical variables such as an alarm type, and numerical variables such as a signal encoding a short video sequence}. 
    
    \item[iv)] Another type of data is \textit{evolving graphs} which consist of graphs whose structure changes over time \cite{latouche2015graphs}. \RemoveForShortVersion{A typical example of an evolving graph is the one which represents the social network of the customers of a telecom provider (i.e. the nodes of the graph) and their interactions through the phone network (i.e. the arcs of the graph). A new customer may appear, leading to the creation of a new \textit{node} in the graph ; or two customers may meet in real life and initiate phone calls resulting in the creation of new \textit{edges} in the graph.} Several types of learning tasks can be considered, such as predicting the next changes in the graph structure, or the classification of parts of the graph (e.g. nodes, arcs, sub-graphs). 
    
    \item[v)] Successive snapshots of \textit{relational data} \cite{dvzeroski2009relational} should be consider to design new ML-EDM approaches. More precisely, relational data consists of a collection of tables having logical connections between them. \RemoveForShortVersion{(e.g. in a relational database system, two tables can be linked together by a foreign key). Generally, raw data stored in information systems can be represented by this type of data, which makes relational data a very widespread data type. 
    For instance, consider that the customers of a company are described in a main table, each row containing the information about a particular customer. The contracts subscribed by the customers can be represented by a secondary table, linked to the main table with a one to many relationship.   
    }
    Like other types, relational data can evolve over time: (\textit{i}) the connections between tables can change; (\textit{ii}) as well as the structure of the tables; (\textit{iii}) or even the values of the information stored in the tables.

    \item[vi)] \textit{Text} is another widespread type of data. \RemoveForShortVersion{There are many application cases where text data is collected over time, and a decision has to be made at a certain moment. For example, this is the case of an e-mail exchange to sell a second-hand car, where the seller realizes after several exchanges that it is most likely a scam. 
    }
    An application example is the moderation of social networking platforms, with early deletion of inappropriate contents and automatic closure of fraudulent accounts (see Section \ref{sec_usecase_social_network}).
    \RemoveForShortVersion{In the Machine-Learning literature, there is little work which considers early decision on texts \cite{he2018time, xia2020state} even though this is likely a future application area of considerable interest, and the development of ML-EDM methods would enable to optimize the time of decision making for text data.} 
\end{enumerate}

%-------------------------------------------------------------------
\bigskip
\noindent
{\bf Challenge \#4: data type agnostic ML-EDM}~\\ 
Ideally, the new developed ML-EDM approaches should be \textit{data type agnostic}, i.e. they should operate for any data type presented above. To do so, a pivotal format needs to be defined in order to learn the triggering strategies in a generic way. For instance, each learning example could be characterized by a series of $T$ predictions indexed by time (corresponding to the output of the learned hypothesis $h({\mathbf x}_t)$ for each time step $t \in [1,T]$), as well as by $\{y_i, {{(s_i,e_i)}}\}_{i=1}^{k_{\mathbf x}}$ the ground truth composed of the true decisions to be made over time for this individual. In the particular case of ECTS, some approaches can easily be adapted to become agnostic to data type \cite{mori2017reliable, mori2019early, achenchabe2021MLj}. 
In contrast, others have been designed to be very specific to time series \cite{xing2009early,xing2011extracting,ghalwash2014utilizing,he2015early}, 
especially with the search of features (e.g. shapelets) occurring early in the time series and helping to discriminate between classes.
More generally, future work in ML-EDM should definitely promote data type agnostic approaches, to allow the use of these techniques in a wide range of application conditions.

%\newpage
%-------------------------------------------------------------------
%-------------------------------------------------------------------
\section{Online Early Decision Making}
\label{stream}
%-------------------------------------------------------------------
%-------------------------------------------------------------------

In the specific case of Early Classification of Time Series (ECTS), an important \textit{limitation} is that the training time series: (\textit{i}) have the same length $T$ ; (\textit{ii}) correspond to different \textit{i.i.d}  individuals ; (\textit{iii}) have a label which characterizes the whole time period of length $T$.
There are obviously applications where this formulation of the problem is relevant \cite{teinemaa2018alarm,alipour2019machine,russwurm2019early,fahrenkrog2019fire,sharma2020novel,loyola2017learning,dachraoui2013early,9207873}, especially in cases where the \textit{start} and \textit{end} of the time series are naturally defined (e.g. a day of trading takes place from 9:30am to 4pm, during the opening hours of the stock exchange). 

The development of \textit{online} ML-EDM approaches could overcome these limitations and enable a new range of applications.
For this purpose, let us consider that 
the input measurements are observed without interruption, in the form of a \textit{data stream} \cite{gama2012survey}. 
In the case of a \textit{classification problem}, an online ML-EDM approach would consist in identifying \textit{chunks} in the input data stream (i.e. fixed time-windows defined by their start and end timestamps) and \textit{categorizing} them according to a predefined set of classes. 
For example, in a predictive maintenance scenario \cite{ran2019survey} such an approach would operate on a continuous basis to detect periods of system malfunction as soon as possible.

\begin{figure}[htbp!]
\centering
\includegraphics[width=0.75\linewidth]{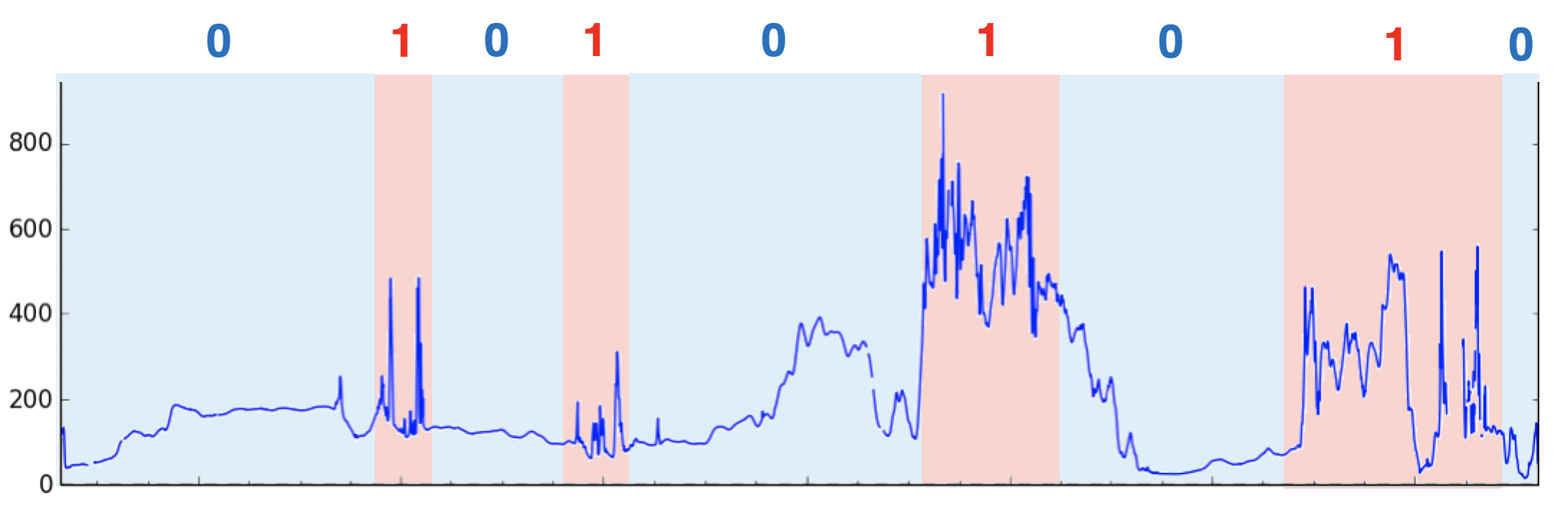}
\caption{Example of a data stream labeled by chunks over a time period}
\label{fig-labelled-stream}
\end{figure}

\bigskip
\noindent
{\bf Challenge \#5: \\ online and early predictions to be located in time}~\\
In the case of a \textit{classification} problem, the training data consist of the measurements observed from the stream during the training period, denoted by ${\mathbf x} \, = \, \langle {x_1}, {x_2}, \ldots, {x_{|{\mathbf x}|}} \rangle$, associated with their labels ${\mathbf y} \, = \, \langle {y_1}, {y_2}, \ldots, {y_{|{\mathbf x}|}} \rangle$.
A \textit{labeled chunk} is formed by 
the consecutive measurements, between the timestamps $t_a$ and $t_b$, if their labels share the same value (i.e. if $ \{y_i\}_{ i \in [t_{a}, t_{b}]} $ is a singleton).
As shown in Figure \ref{fig-labelled-stream}, the data stream defined over the training period is labeled by chunks of variable size.
For example, these chunks could represent the periods of failure and nominal operation in a predictive maintenance scenario. 
During the deployment phase, the model is applied online on a data stream whose measurements are observed progressively over time. This model is expected to provide \textit{predictions located in time}, since it needs to predict the \textit{beginning} and the \textit{end} of each chunk, associated with the predicted \textit{class} which characterizes the state of the system during this chunk.

\bigskip
\noindent
{\bf Challenge \#6: online accuracy vs. earliness trade-off}~\\
Designing \textit{online} ML-EDM approaches requires redefining the accuracy vs. earliness trade-off for online decisions. 
The main issue is that a data stream is of indeterminate length: (\textit{i}) its \textit{beginning} may be too old to be considered explicitly, or can even be indeterminate ; (\textit{ii}) its \textit{end} is never reached, since it is constantly postponed by the new measurements which arrive.  
In the particular case of ECTS, it is precisely the fact that the input series has a maximum length $T$, known in advance, that leads to force triggering the decision when the current time $t$ becomes close to the \textit{deadline} $T$.

%\vspace{2mm}

\bigskip
\noindent
{\bf Challenge \#7: \\management of non-stationarity in ML-EDM}~\\
It is not always realistic to assume stationarity of the data. In practice, data collected from a stream may suffer from several types of drifts: 
(\textit{i}) the distribution of the measurements within the sliding window ${\mathbf x}_t$ can vary over time, this is called \textit{covariate-shift} \cite{quinonero2009dataset};
(\textit{ii})  the prior distribution of the classes $P(y)$ can be subject to such drifts;
(\textit{iii}) and the concept to be learned $P(y|{\mathbf x})$ can also change when \textit{concept drift} occurs \cite{gama2010knowledge}.

To manage these non-stationarities, a first family of approaches maintains a decision model trained 
using a sliding window of most recent examples. This is a blind approach, in the sense that there is no explicit drift detection.
The main problem is deciding the appropriate window size.

A second family of approaches, explicitly \textit{detects} the drifts \cite{GamaZBPB14,lemaire2014survey} 
and triggers actions when necessary, such as re-training the model from scratch, 
or using a collection of models in the case of ensembles. 
In this case, detecting concept drift can be considered as similar to the anomaly detection problem, and ML-EDM approaches could be used to tackle it in future work.
A popular idea is to train the decision model using a growing window while data is stationary, 
and shrink the window when a drift is detected.
These kinds of approaches can easily be adapted to \textit{online} ML-EDM, 
since they decouple model training and non-stationarity management. % 

In the case of incremental concept drift, a third family of approaches consists in continuously adapting the model 
by training it online from recent data. This kind of \textit{adaptive} approach is much more challenging to adapt to 
\textit{online} ML-EDM. 
Indeed, as in ML-EDM problems (see Figure \ref{fig:cost-origin}), two kinds of models are used: (\textit{i}) the \textit{predictive model(s)}, which can categorize the input data stream at any time ; (\textit{ii}) the \textit{triggering strategy } which makes the decisions at the appropriate time. 
The main challenge in developing adaptive drift management methods for the \textit{online} ML-EDM problem is that the parameters of the \textit{predictive models} and of the \textit{triggering strategy } must be updated jointly. 
These two kinds of models are highly dependent: updating the parameters of one has an impact on the optimal parameters of the other. 

By contrast, in standard ML-EDM approaches which operate in batch mode, the parameters of the predictive models are first optimized, and then the parameters of the triggering strategy are optimized in turn given the parameters of the classifiers (see paragraph B in Section \ref{definition-ML-EDM}).
This two-step Machine Learning scheme is definitely not valid for managing drift online \cite{Krempl2014OCD}. 
Adaptive drift management for the \textit{online} ML-EDM problem has not yet been addressed in the literature and constitutes an interesting research direction. 
In drift detection systems, there is a trade-off between fast detection and the number of false alarms. 
Moreover, in problems where the target (e.g. the labels) is not always available or available with a delay requires unsupervised or semi-supervised drift detection mechanisms.
The ML-EDM framework, improving the compromise between earliness and accuracy,  can provide new approaches for drift detection.

%-------------------------------------------------------------------
%-------------------------------------------------------------------
\section{Revocable decisions}
\label{revocable_decision}
%-------------------------------------------------------------------
%-------------------------------------------------------------------

In many situations, one can take a decision and then decide to change it after some new pieces of information become available. 
The change may be burdensome but nevertheless justified because it seems likely to lead to a much better  outcome. 
This can be the case when a doctor revises what now seems a misdiagnosis. 

Similarly, ML-EDM should be \textit{extended} to consider such a revocation mechanism.
In the classical ML-EDM problem as described in Section \ref{definition-ML-EDM}, a prediction $h({\mathbf x}_{\hat{t}})$ cannot be changed once the decision is triggered at time $\hat{t} \leq T$. The cost of such an \textit{irrevocable} decision is given by the loss function described by Equation \ref{eq:double-loss-function}. Whereas, the extension of ML-EDM to \textit{revocable} decisions \cite{achenchabe2021earlyrev} allows a prediction to be modified several times before the deadline $T$.  On the one hand, the revocation of a decision generates a higher delay cost $\mathcal{L}_{delay}$, as well as a cost of changing the decision $\mathcal{L}_{revoke}$. On the other hand, new data observed in the meantime provide information that makes the prediction more reliable, thus tending to decrease the misclassification cost $\mathcal{L}_{prediction}$. Ultimately, the main issue is to identify the appropriate decisions to revoke, in order to minimize the global cost, given by Equation \ref{eq:loss-function-2}.
 
Such an extension to revocable decisions could be of great interest: (\textit{i}) in applications where the cost of changing decisions is low, i.e. the DAGs associated with each possible decision share reusable tasks (see Section \ref{decision_costs}) ; (\textit{ii}) in applications involving online early decision making (see Section \ref{stream}).
There are many use cases where the need to \textit{revoke} decisions appears clearly.
For instance, the emergency stop system of an autonomous car brakes as soon as an obstacle is suspected on the highway, and  releases the brake when it realizes, as it gets closer, that the suspected obstacle is a false positive (e.g. a dark spot on the road).   

\bigskip
Developing ML-EDM approaches capable of appropriately revoking its decisions involves solving the two following challenges:

%-------------------------------------------------------------------
\bigskip
\noindent
{\bf Challenge \#8: \\reactivity \textit{vs.} stability dilemma for revocable decisions}~\\
The first issue is to ensure that a decision change is driven by the information provided by the recently acquired measurements, and not caused by the inability of the system to produce a stable decision over time.
This problem is not trivial. 
On the one hand, the system needs to be reactive by changing its decision promptly when necessary.  
On the other hand, the system is required to provide stable decisions over time by avoiding excessively frequent and undue changes. 
Thus, a trade-off exists between the \textit{reactivity} of the system and its \textit{stability} over time. 
One way to formalize this trade-off is to associate a cost to \textit{decision changes}, as it is proposed in part (iii) of Equation \ref{eq:loss-function-2}.  
To our knowledge, only one approach uses such a cost of decision change \cite{achenchabe2021earlyrev}, in order to penalize revocation of too many decisions.
The \textit{reactivity vs. stability} dilemma of revocable decisions is understudied in the literature, and it would be interesting for the scientific community to work on this question.

%-------------------------------------------------------------------
\bigskip
\noindent
{\bf Challenge \#9: extending non-myopia to revocation risk}~\\
Non-myopic ML-EDM approaches are capable of estimating the information gain that will be provided by future measurements, based on the currently visible ones. 
In other words, these approaches are able to predict the reliability improvement of a decision in the future. 
Thus, a decision is triggered when the expected gain in miss-classification cost at the next time steps does not compensate the cost of delaying the decision \cite{achenchabe2021MLj}. 
In the case of revocable decisions, an important challenge is to estimate the future information gain by taking into account the \textit{risk of revocation} itself. 
Specifically, a decision that will probably be revoked afterward should be delayed due to this risk.
Conversely, a decision which promises to be sustainable should be anticipated. 
Designing \textit{non-myopic to revocation risk} approaches could be an important step forward to (\textit{i}) optimize the first trigger moment, and (\textit{ii}) reduce the number of undue decision changes. 
The approach proposed in \cite{achenchabe2021earlyrev} constitutes a first step in this direction, by assigning a cost to decision changes and considering it in the expectation of future costs. 
To the best of our knowledge, this is the only approach which provides this interesting property. It is not clear whether alternative methods are possible. This is an interesting topic for further studies by the scientific community.

%-------------------------------------------------------------------

\section{Origin of the decision costs} 
\label{cost-origin2}

\begin{figure}[htbp!]
\centering
\includegraphics[width=0.7\linewidth]{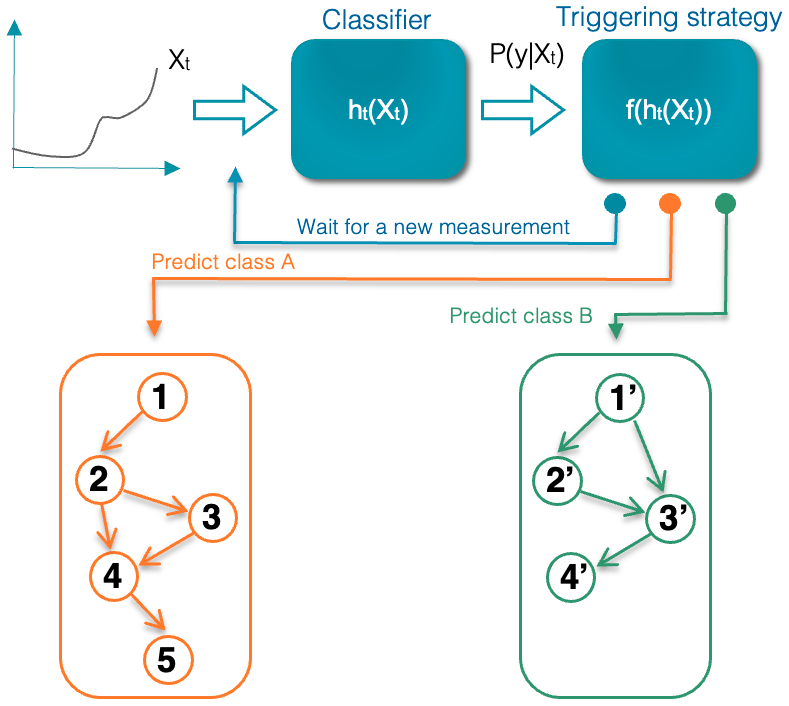}
\caption{DAGs of tasks to be performed after the triggering of a decision.}
\label{fig:cost-origin-2}
\end{figure}

The origin of the delay cost is studied in Section \ref{cost-origin}, however it is necessary to further specify the operating scenario in order to understand the other decision costs involved in ML-EDM. 
Figure \ref{fig:cost-origin-2} describes a binary ECTS problem, where the actions to be performed depend on the predicted class and are described by two \textit{Directed Acyclic Graphs} (DAG). These DAGs characterize the sequence and the relationships between the unit tasks which compose them (e.g. task 1 must be completed before starting task 2). Here, the DAGs of tasks are \textit{fixed}, they do not depend on the decision time.

%\bigskip
The total cost of a decision can be decomposed by: 
\begin{enumerate} [label=(\roman*)]
    \item the \textit{delay} cost, denoted by $\mathcal{L}_{delay}$, which reflects the need to execute the DAG of actions corresponding to the new decision in a constrained time, and in a parallel way {(already detailed in Section \ref{cost-origin})};
    
    \item  the \textit{decision} cost, which corresponds to the consequences of a bad decision, or the gains of a good decision  (denoted by $\mathcal{L}_{prediction}$).
    
    \item  the \textit{revocation} cost, which is the cumulative cost of the mistakenly performed tasks belonging to the DAG of previously made bad decisions, and which are not reusable for the new decision (denoted by $\mathcal{L}_{revoke}$) ; 
    
\end{enumerate}

When expressed in the same unit, these different types of costs can be summed up in order to reflect the quality of the decisions made, and their timing. Thus, Equation \ref{eq:loss-function-ECTS} becomes:

\begin{equation}
\begin{split}
    \mathcal{L}(h({\mathbf x}_t), t, y) &= \overbrace{\mathcal{L}_{delay}(t)}^{(i)} + \overbrace{\mathcal{L}_{prediction} \left( h(\mathbf{x}_t),y \right)}^{(ii)}   \\
    &+ \underbrace{\mathcal{L}_{revoke} \left (h(\mathbf{x}_t)|\{ ( h({\mathbf x}_{\hat{t_{i}}}), \hat{t_{i}} ) \}_{i \in [1,D^{\mathbf x}_t]} \right )}_{(iii)}
\end{split}    
\label{eq:loss-function-2}    
\end{equation}

\noindent
where $\{ ( h({\mathbf x}_{\hat{t_{i}}}), \hat{t_{i}} ) \}_{i \in [1,D^{\mathbf x}_t]}$ represents the sequence of the previously made decisions and their associated triggering time, with $\hat{t_{i}} < t, \forall i \in [1, D^{\mathbf x}_t]$.

\bigskip
\noindent
Term (\textit{ii}): Taking into account the \textit{decision} cost is a very common feature in the literature, particularly in the field of cost-sensitve learning \cite{elkan2001foundations}.
These techniques take as input a function $\mathcal{L}_{prediction}(\hat{y}|y) : {\cal Y} \times {\cal Y} \rightarrow \mathbb{R}$ which defines the cost of predicting $\hat{y}$ when the true class is $y$. 
The aim is to learn a classifier which minimizes these costs on new data.

\bigskip
\noindent
Term (\textit{iii}): By contrast, the study of the \textit{revocation} cost is very limited in the literature. To our knowledge, \cite{achenchabe2021earlyrev} is the only one article article that considers this problem, and this work shows that assigning a cost to decision changes is a first lead to manage the \textit{reactivity vs. stability dilemma}, and to design \textit{non-myopic to revocation risk} approaches (i.e. discussed later in challenges {\small \#}8 and {\small \#}9). 
The origin of this cost can be explained in the light of the tasks to be performed once a decision is triggered (see Figure \ref{fig:cost-origin-2}). 
For instance, let us consider the first decision noted by $(A,\hat{t_1})$, in which the system predicts at time $\hat{t_1}$ that the input time series belongs to the class $A$.  
This decision is then revoked in favor of a new decision $(B,\hat{t_2})$.  
The cost of changing this decision, denoted by $\mathcal{L}_{revoke}((B,\hat{t_2})|(A,\hat{t_1}))$, can be defined as the cost of the actions already performed between $\hat{t_1}$ and $\hat{t_2}$ which turn out to be useless for the new decision, i.e. which cannot be reused in the DAG of tasks corresponding to the new predicted class $B$. 
In order to define the costs of decision changes, it is necessary to identify the \textit{common tasks} between the DAGs of the different classes and to evaluate their execution time. 
In addition, the entire sequence of the past decisions must be taken into account to identify the already completed tasks which are now useful for the achievement of the current DAG of tasks. 
For instance, the cost $\mathcal{L}_{revoke}((A,\hat{t_3})|\{(A,\hat{t_1}),(B,\hat{t_2})\})$ can be reduced by the tasks executed between $\hat{t_1}$ and $\hat{t_2}$, if these tasks are \textit{not perishable}, i.e. the results %already obtained 
are identical to those that would be obtained by re-executing these tasks at $\hat{t_3}$.

%%%%%%%%%%%%

\begin{figure}[htbp!]
\centering
\includegraphics[width=0.7\linewidth]{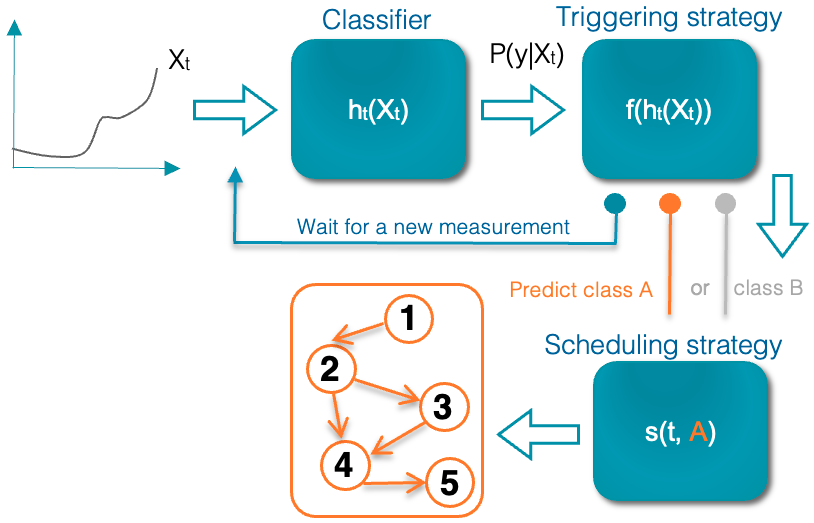}
\caption{DAG of tasks to be performed after the triggering of a decision, generated by a scheduling strategy.}
\label{fig:cost-origin-3}
\end{figure}

\bigskip
\noindent
{\bf Challenge \#10: \\scheduling strategy and time-dependent decision costs}~\\ 
In this paper, the DAGs of tasks are supposed to be \textit{fixed}, i.e. not depending on the decision time. 
However, a more general problem could be considered (see Figure \ref{fig:cost-origin-3}) where the DAGs of tasks are generated by a \textit{scheduling strategy} depending on: (\textit{i}) the decision made ; (\textit{ii})  and the decision time.
Such a scheduling strategy is useful in applications where the actions to be performed after a decision can be \textit{adapted} to a time budget available to perform them. 
Two situations may occur:  
(\textit{i}) ideally, a decision is triggered early enough to allow the scheduling strategy to generate a \textit{complete} DAG of tasks which is optimal given the decision made (as in Figure \ref{fig:cost-origin-2}) ;  
(\textit{ii}) on the contrary, in the case of a too late decision, the scheduling strategy needs to build the DAG so that it can be achieved in the remaining time (e.g. by parallelizing some tasks, by changing or removing some of them).
For instance, when flying an airplane, the tasks to be performed for an emergency landing are not the same as for a normal landing, and there is a range of situations with different emergency level, and therefore corresponding to different time budgets.

Such a time-dependent scheduling strategy radically transforms the ML-EDM problem and the way it can be formulated. In particular, the \textit{triggering} and \textit{scheduling} strategies become mutually dependent:

\begin{enumerate}
    \item \textit{Decision costs depend on the generated DAG of tasks:} all the previously discussed costs result from the structure of the DAG to be performed conditionally to the decision made: (\textit{i}) the relationships between the tasks ; (\textit{ii}) their execution time ; (\textit{iii}) the conditions of their reuse when they are common to several DAGs. Since the structure of the DAG to be performed now depends on the decision time, the decision costs can no longer be considered as fixed, and they are available only after scheduling.
    
    \item \textit{The optimal decision time depends on the cost values:} on the other hand, the triggering strategy aims to optimize the decision time based on the cost values. As described in Equation \ref{eq_learn_optimal_strategy}, the triggering strategy is learned by minimizing the empirical risk, which is itself estimated using a loss function based on the costs. 
    
\end{enumerate}

This mutual dependency between the triggering and the scheduling strategies has strong impacts on the ML-EDM problem. 
In particular, the optimal decision time $t^*$ described in Equation \ref{eq:trigger_function} must be redefined as a \textit{fixed point}, i.e. the function to be optimized takes the optimal solution as an input parameter, such that $t^* = \argmin_{t^* \in [1, T]} \mathcal{L}(h({\mathbf x}_{t^*}), t^*, y)$. This leads to a much more difficult class of optimization problems, for which the simple existence of a solution is difficult to ensure.  

Finding an optimal triggering strategy when the scheduling strategy is itself time-dependent makes  ML-EDM a quite difficult challenge as the scheduling strategy is only known through its interactions with the triggering strategy.
In this case, Reinforcement Learning seems to be a possible option to address the problem. The scheduling strategy could then be considered as part of the environment, and a contributor to the reward signal by determining the decision costs for each decision taken at a particular time. However, this line of attack remains to be investigated in order to assess its merit.

In many applications, fortunately, the implementation of a scheduling strategy is much simpler, especially when the variation of decision costs over time is known in advance (or modeled, and thus are partially known). 
We will place ourselves under this assumption in the rest of this paper. The preceding remarks are reminders that if considered in all its complexity, ML-EDM becomes a very difficult problem. As we will see below, addressing the case where the costs are assumed to be time dependent but with a known form, already offers interesting challenges and corresponds to a variety of applications.

\section{Overview on challenges}
\label{sec_overview}

This section provides an overview of the previously presented challenges, indicating references which address part of these challenges (see the second column of Table \ref{table:overview_methods}), and summarizing the main prospects for applications in the short and long term (see the last column of Table \ref{table:overview_methods}).
Table \ref{table:overview_methods} organizes the proposed challenges by category, using colors to identify: (\textit{i}) those related to changing the learning task ; (\textit{ii}) those related to online ML-EDM ; (\textit{iii}) and those related to revocable decisions. 

%\newpage

% -------------------------------------------------------------------

%\begin{center}
\begin{table*}[ht]
\scriptsize
\begin{tabularx}{\textwidth}{| L{0.42} | L{0.08} | L{0.50} |}
 \hline 
 {\bf ML-EDM challenges} 	&  {\bf SOTA} 	& \textbf{Main application perspectives}  \\
 \hline
 \hline
 \cellcolor{my_blue_lite} 
 {\bf \leavevmode ~\newline \#1 {\tiny (Section \ref{learning_tasks})} \leavevmode \newline Extending non-myopia to \leavevmode \newline unsupervised approaches} 		&	 \leavevmode \newline ~ \newline{\large } 	&  \leavevmode ~\newline {\bf } In anomaly detection applications, anticipate the deviation of an observed individual from a normal behavior.    \newline	\\
 \hline
 \cellcolor{my_blue_lite} 
 {\bf \leavevmode ~\newline \#2 {\tiny (Section \ref{learning_tasks})} \leavevmode \newline Addressing other supervised learning tasks} 		& \leavevmode \newline ~ \newline{\large }	 &  \leavevmode ~\newline {\bf } Adapt ECTS approaches to extrinsic regression problems.\leavevmode \newline ~\newline{\bf } Develop forecasting methods whose prediction horizon can adapt.    \newline  	\\
 \hline
 \cellcolor{my_blue_lite} 
 {\bf \leavevmode ~\newline \#3 {\tiny (Section \ref{learning_tasks})} \leavevmode \newline Early weakly supervised \leavevmode \newline learning (WSL) \leavevmode \newline} 		& \leavevmode \newline ~ \newline{\large }	&  \leavevmode ~\newline {\bf } Adapt ECTS approaches to the different WSL classification scenarios.  \newline 	\\
 \hline
 {\bf \leavevmode ~\newline \#4 {\tiny (Section \ref{data_types})} \leavevmode\newline Data type agnostic ML-EDM} 		&	 \leavevmode ~\newline \cite{dachraoui2015early,achenchabe2021MLj,mori2015early, mori2017early} 	&  \leavevmode ~\newline {\bf } Identify agnostic approaches in the literature and promote this feature. \leavevmode \newline ~ \newline{\bf } Define a pivotal format allowing to develop an ML-EDM library.    \newline 	\\
 \hline
 \cellcolor{my_green_lite} 
 {\bf \leavevmode ~\newline \#5 {\tiny (Section \ref{stream})} \leavevmode \newline Online predictions to be \leavevmode \newline located in time} 		&	 \leavevmode \newline ~ \newline{\large } 	&  \leavevmode ~\newline {\bf } Applications where the arrival of an event (e.g. a failure) must be predicted in advance, as well as its duration. 	 \newline  	 \\
 \hline
 \cellcolor{my_green_lite} 
 {\bf \leavevmode ~\newline \#6 {\tiny (Section \ref{stream})} \leavevmode \newline Online accuracy vs. earliness trade-off \leavevmode \newline} 		& \leavevmode ~\newline	\cite{achenchabe2022ECOTS} &  \leavevmode ~\newline {\bf } Optimize decision time in online predictive maintenance applications.   \newline 	\\
 \hline
 \cellcolor{my_green_lite} 
 {\bf \leavevmode ~\newline \#7 {\tiny (Section \ref{stream})} \leavevmode \newline Management of \leavevmode \newline non-stationarity in ML-EDM \leavevmode \newline} 		&	 \leavevmode \newline ~ \newline{\large} 	&  \leavevmode ~\newline {\bf } Properly manage the potentially long life of ML-EDM models.   \newline 	 \\
 \hline
 \cellcolor{my_purple_lite} 
 {\bf \leavevmode ~\newline \#8 {\tiny (Section \ref{revocable_decision})} \leavevmode \newline Reactivity vs. stability dilemma for revocable decisions\leavevmode \newline} 		&	 \leavevmode ~\newline \cite{achenchabe2021earlyrev} 	&  \leavevmode ~\newline {\bf } Applications where undue and excessive decision changes must be avoided.    \newline 	\\
 \hline
 \cellcolor{my_purple_lite} 
 {\bf \leavevmode ~\newline \#9 {\tiny (Section \ref{revocable_decision})} \leavevmode \newline Non-myopia to revocation risk} 		&	 \leavevmode ~\newline \cite{achenchabe2021earlyrev} 	&  \leavevmode ~\newline {\bf } Applications where it is necessary to delay decisions which are likely to be changed later.   \newline 	\\
 \hline
 {\bf \leavevmode ~\newline \#10 {\tiny (Section \ref{decision_costs})} \leavevmode \newline scheduling strategy and time-dependent decision costs}		& \leavevmode \newline ~ \newline{\large }	&  \leavevmode ~\newline {\bf } Applications where the variation of the decision costs over time is known or can be modeled. \leavevmode \newline ~ \newline{\bf } Applications where the scheduling strategy is only known through its interactions with the triggering strategy.  \newline 	\\
\hline
\end{tabularx}
\vspace{1mm}
\caption{Overview of the proposed challenges by category: in \textcolor{blue}{\bf blue} those related to the \textit{learning task}, in \textcolor{ForestGreen}{\bf green} those related to \textit{online} ML-EDM, in \textcolor{Purple}{\bf purple} those related to \textit{revoking decisions}, and in white the others. \\
}
\label{table:overview_methods}
\end{table*}
%\end{center}
%\end{landscape}

%-------------------------------------------------------------------
%-------------------------------------------------------------------
\section{Usecases}
\label{usecases}
%-------------------------------------------------------------------
%-------------------------------------------------------------------

ML-EDM approaches can be applied to a wide range of applications, such as cyber security \cite{Tommaso2021}, medicine \cite{khoshnevisan2021}, surgery \cite{Jackson2021}. This section develops some key use cases and identifies possible advances in near future, if the proposed challenges are met.

\subsection{Early classification of fetal heart rates}
\label{section-fetal-heart}
There are no precise figures on the number of deaths in childbirth due to poor oxygenation. According to the Portuguese Directorate-General for Health, the number of children who died due to hypoxia in 2013 was 192 fetuses.
This is a critical example where 
making informed early decisions is critical\RemoveForShortVersion{, literally meaning a difference between life and death}.
Cardiotocography 
techniques are used to assess fetal well-being through continuous monitoring of fetal heart rate and uterine contractions \citep{Luzietti99}. 
\RemoveForShortVersion{
Fetal well-being results from the normal functioning of the transfer of maternal blood to the placenta and, through its proper functioning, the transfer of oxygen present in maternal blood to fetal blood \citep{Neilson13}.}
Labor is a potentially threatening situation to fetal well-being, as strong uterine contractions stop the flow of maternal blood to the placenta, compromising fetal oxygenation \citep{Neilson13}. 
\RemoveForShortVersion{
Hypoxia, resulting from lack of oxygen, represents a large part of unsuccessful deliveries.  
In addition, more than 50\% of deliveries with poor outcome are caused by failure to recognize fetal heart rate patterns \citep{Chinnasamy13}.}
 
In this field, ML-EDM techniques could be of great help to detect the early warning signs of complications during childbirth. This application can be addressed as a ECTS problem, as  
a fetal heart rate signal constitutes a time series. 
\RemoveForShortVersion{The goal there is twofold: (\textit{i}) classifying the birth outcome before having access to complete time-series ; (\textit{ii}) optimizing the triggering moment, when this prediction is made.} 
The extension of ECTS techniques to revocable decisions would be very relevant (see challenges \#8 and \#9) allowing for active monitoring of the children's well-being on a continuous basis, until delivery. 
In addition, two particular aspects need to be taken into account in developing an efficient approach: (\textit{i})
the prediction cost $\mathcal{L}_{prediction}$ is highly asymmetrical since a false negative can mean the death of the baby or the mother;
(\textit{ii}) the deadline $T$ which represents the moment of delivery is uncertain and varying. Thus, the deadline $T$ corresponds to the occurrence of an event (i.e. the birth ) which can be modeled as random variable as in \cite{kochenderfer2015decision, frazier2007sequential}.

\subsection{Digital twin in production systems}

Digital Twin (DT) is an important active concept in the area of Industry 4.0. With the development of low cost sensors and efficient IoT communication facilities, almost all production systems are now equipped with several sensors enabling real time monitoring and helping in decisions about maintenance, or when failures occur. 
\RemoveForShortVersion{Production systems with sensors coupled with computers are also called cyber-physical systems (CBS).}
In this section, we consider digital twins (DT) of cyber-physical systems (CBS) which are in operation.

%\medskip
%In \cite{Fuller2020}, it is reported that 
The main digital twin applications \cite{Fuller2020} are related to smart cities, manufacturing, healthcare and industry. 
The role of the DT is thus to use the data streams coming from the sensors of the CBS in order to constantly calibrate simulation models of different components of the system.
Indeed, this offers 
several opportunities, namely (1) detection of anomalies when the system deviates from the simulation model ; (2) diagnostic of dysfunctions when they occur ; (3) exploration of different scenarios for system evolution in case of dysfunction ; (4) recommendation for repair actions. 
\RemoveForShortVersion{
These approaches were, for instance, well illustrated in the European H2020 project MAYA (see \url{http://www.maya-euproject.com/}).
}

%\medskip
Effective maintenance management methods are vital, and industries seek to minimize the number of operational failures\RemoveForShortVersion{, reduce their operational costs, and increase their productivity}.
\RemoveForShortVersion{In this context,} The availability of large volume of data coming from sensors of a  CBS makes the use of Machine Learning techniques, supervised or unsupervised, very appealing. 
Typical unsupervised ML approaches are related to anomaly detection \cite{Ruff2021} where an alarm should be triggered when the behavior of the CBS differs from normal running. Typical supervised ML approaches in the context of manufacturing and industry are related to predictive maintenance \cite{CARVALHO2019106024,ran2019survey}\RemoveForShortVersion{, where classification models are used to predict  categories that correspond to possible failures}. 
Predictive Maintenance (PdM) is a data-driven approach that emerged in Industry 4.0\RemoveForShortVersion{ as a prominent field of research}. It uses statistical analysis, Machine Learning (ML) models\RemoveForShortVersion{, and Deep Learning (DL) solutions} for modeling complex systems behavior, identifying trends and predicting failures. 

\RemoveForShortVersion{
In both anomaly detection and predictive maintenance, ML models need to be updated continuously with incoming new data, and their outputs are used by operators for decision making. Consequently, DT's using machine learning models can benefit from early decisions and ML-EDM techniques. The \textbf{multiple early decision to be located in time} concept introduced in Section \ref{definition-ML-EDM} is particularly relevant in this context, since maintenance operations often imply several periods of time.}

%\smallskip
We review below some challenges of the paper in light of this domain.
%
%\smallskip
%\RemoveForShortVersion{
%As for challenge \#10, the definition of prediction and delay costs is rather easy, since they can be obtained directly from experts managing the industrial asset. 
%For instance, delaying a maintenance operation is directly associated with costs of a breakdown in the manufacturing process.} 
Challenge \#1 (extending non-myopia to unsupervised approaches) is relevant, since an efficient anomaly detection system requires unsupervised approaches which can be combined with physics-based simulations of the different components. Challenge \#2  (other supervised tasks) is also appropriate since both classification and regression problems appear (e.g. breakdown occurrence, prediction of energy consumption).
\RemoveForShortVersion{
Regarding challenge \#3 (weakly supervised learning), the main problem is not that training data is of poor quality but that interesting data (e.g. failures) are often rare, leading to the need for data augmentation using simulation.} \RemoveForShortVersion{Of course} Challenge \#4 (data type agnostic) is especially relevant for DT's, since a system is always composed of several heterogeneous components\RemoveForShortVersion{with many sensors generating data of heterogeneous type, for instance tabular data, multivariate time series, images, audio signals such vibrations, possibly videos, ...)}. In this situation, the update of one component or one or several sensors would be much easier and cheaper if ML-EDM were data type agnostic. DT's operating at a system level leads to complex prediction models and complex decisions since the different components operate differently but in interaction (cf. challenge \#5).
The ability to manage non-stationarity (cf. challenge \#7) is obviously central in DT's: aging and wearing of equipment lead to covariate and concept drifts which must be taken into account. \RemoveForShortVersion{Moreover, prediction models may need to be recalibrated after maintenance operations.}  

\RemoveForShortVersion{
\subsubsection*{Scenario exploration in manufacturing digital twins}
\label{sec_scenario_exploration}

As mentioned above, one interesting characteristic of Digital Twins (DT's) is their ability to anticipate possible evolution of the system using current observation and running simulation models. For instance in \cite{Gabor2016}, a software architecture framework is defined which enables information exchanges between the cyber-physical system and simulation models which are part of the DT. This enables simulations of possible evolution of the CBS if such simulations can be operated in real-time. In \cite{Lugaresi2018}, a review of Real-Time Simulation (RTS) models is proposed and describes several approaches for RTS, usually based on Discrete Event Simulation (DES). As reported in this paper, there are still research challenges in RTS for manufacturing (data management, adaptability, model generation, validation, reactiveness) but some solutions exist. In \cite{Lugaresi2019}, a LEGO toy demonstrator has been developed to prove this concept. 

Consequently, one can consider that in manufacturing, Digital Twins will soon have the ability to run in real-time simulations of the evolution of the system, based on Discrete Event Simulation models. These simulations provide different possible outcomes of the system with associated probabilities: typically what an ML-EDM expects as input in a \textit{non-myopic} decision perspective. This case is closely related to the problems addressed in challenge \#10 (scheduling strategy and time-dependent decision costs), since the time needed to run simulations is similar to the time budget of a scheduling strategy (cf. Figure \ref{fig:cost-origin-3}).}

\RemoveForShortVersion{
\subsection{Predictive Maintenance of Metro Trains}
\label{sec_predictive_maintenance}

Advances in networking, machine learning, data analytics, and robotics are allowing vast improvements on industrial processes. Predictive maintenance, in particular, is one technique with high impact in today's industry.
Over the years different maintenance strategies have been developed. Three main approaches can be identified:
({\textit{i}}) \textit{Corrective} maintenance: when an equipment is run until failure.
  This simplest of method almost always leads to high (unexpected) downtime and thus potentially to  critical situations that entail great costs for companies.
({\textit{ii}}) \textit{Preventive} maintenance: It is based on planning regular replacement of components and/or equipment. Historical failure data and/or the data provided by the equipment manufacturer is used.
  Although this approach prevents unexpected shutdown, it usually entails unnecessary additional costs and an increased unexploited lifetime.
({\textit{iii}}) \textit{Predictive} maintenance: It uses direct monitoring of the mechanical condition and other parameters that can determine the operating conditions over time in order to accurately predict the arrival of a breakdown. There are now tools that can process real-time data acquired from different equipment parts to predict any sign of failure.

%\bigskip
Data-driven predictive maintenance (PdM) monitors the mechanical conditions or other health indicators of the equipment, and uses advanced statistical or Machine Learning methods to detect operating patterns and  dynamically identify operating conditions.

%\bigskip
Predictive maintenance of metro trains offers a lot of opportunities for ML-EDM to improve the quality of service of public transportation. Clearly, anomaly detection can be addressed using unsupervised techniques, which in turn implies challenge \#1. The data collected on the operations of metro trains can be of various types: multi-variate series, feedback from the drivers or the maintenance agents, and so on, which relate to challenge \#4. Of course, challenges \#5 and \#6 on online predictions with location in time are implied as well. Finally, even though once an alert has been raised and predictive operations have been scheduled, it is rare that the decisions can be revoked, this is nonetheless an issue that can be considered when further data may lead to a reassessment of the situation (see challenge \#8 and \#9). 
}

\subsection{Social networks: societal and psychological risks}
\label{sec_usecase_social_network}

Online social networking platforms are more popular than ever. 
\RemoveForShortVersion{They are now used daily and have become an important part of our lives.} 
They radically transform the way we communicate with each other. However, this transformation comes with many problems on both sides, for users and platforms\RemoveForShortVersion{ alike}. 

For example, \textit{Fake news} spread widely during the covid pandemic\RemoveForShortVersion{, which had an impact on the spread of the virus itself}.
\cite{ajao2019sentiment} tackled this problem as a binary classification problem where classes are ``fake" and ``real" news\RemoveForShortVersion{, by using fact checking techniques}. \textit{Fake accounts} are also considered a major problem \RemoveForShortVersion{for these platforms}, as they are among the main culprits in spreading false information. For instance,  \cite{elyusufi2019social, fire2014friend, singh2018detection, aydin2018detection} use Machine Learning techniques to detect these fake account based on interactions between users. 
Fake accounts can also be used for harassment and \RemoveForShortVersion{propagating hate speech, which} can induce major psychological risks \cite{watanabe2018hate}.  The detection of depression and risk of suicide has been addressed using Machine Learning techniques in \cite{castillo2020suicide, islam2018depression}.

%\medskip
Decisions taken by Machine Learning models to prevent such \RemoveForShortVersion{societal and psychological} risks on social networks are clearly \textit{time-sensitive}:  

\begin{itemize}

    \item Fake news must be detected as early as possible to limit its spread\RemoveForShortVersion{, and thus its harmful consequences on society}. \RemoveForShortVersion{For example,} \cite{zhou2020fake} focuses on early detection of fake news from the press before it is expressed on social media\RemoveForShortVersion{, and \cite{liu2020fned} proposes a deep learning model that achieves an accuracy of more than 90\% within 5 minutes of the beginning of the propagation of a fake news and before it is retweeted 50 times}.
 
    \item The early detection of \textit{fake users} has also been studied in recent work. \RemoveForShortVersion{For example,} \cite{breuer2020friend} proposes a graph-based approach which uses network connectivity 
    to detect fake users\RemoveForShortVersion{ as early as possible}. 
    
    \item \RemoveForShortVersion{Likewise,} Detecting as early as possible \textit{depressed} \RemoveForShortVersion{or potential suicidal users} is very critical for prevention. This problem has also been addressed under the perspective of early classification in \cite{leiva2017towards}. 
    
\end{itemize}

The development of the ML-EDM domain is an opportunity to go further in these applications.  
In particular, it would be very useful to develop unsupervised and weakly supervised ML-EDM approaches (see challenges \#1 and \#3). In this application area, ground truth is often unavailable or corrupted. 
\RemoveForShortVersion{Typically, users rarely declare on social networks their depressive state or their suicidal thoughts ; and if they do, this information is not reliable due to obvious social biases.} 
\RemoveForShortVersion{In the case of social networks,} Training data is very complex and consists of multiple sources: streams of texts, a large graph evolving over time etc. Therefore, it would  be particularly beneficial to develop ML-EDM approaches which are agnostic to data types (see challenge \#4).  
\RemoveForShortVersion{In the scenario where the user's state is monitored in a streaming fashion, and where the goal is to identify their state changes as soon as possible, the development of online ML-EDM approaches seems to be very desirable (see challenges \#8 and \#9).}

\subsection{Autonomous vehicle}

An autonomous vehicle is defined in \cite{thrun2010} as capable of sensing its environment and navigating safely without human input. 
\RemoveForShortVersion{
In order to achieve this ambitious goal, it is necessary to combine advanced technologies from many fields such as: (\textit{i}) electronics and sensors ; (\textit{ii}) software engineering ; (\textit{iii}) telecommunication ; (\textit{iv}) computer security ; (\textit{v}) information processing. 
}
Five levels of vehicle automation have been defined \cite{milakis2017} as intermediate goals toward full automation\RemoveForShortVersion{: in \textit{level 1}, most functions are controlled by the driver ; in \textit{level 2}, at least one driver assistance system is implemented ; in \textit{level 3}, the driver is able to delegate safety critical functions to the vehicle ; in \textit{level 4}, the vehicle is fully autonomous, but not in all driving scenarios ; in \textit{level 5}, the vehicle is fully autonomous, with performance equal to that of a human driver in all driving scenarios}.
The development of a fully autonomous vehicle (levels 4 or 5) requires a complex software architecture, which operates numerous functional components \cite{serban2018}. 
More precisely, three classes of components have been identified, corresponding to different levels of control: 

\begin{enumerate}
    
    \item[i)] \textit{Operational} components, which implement basic vehicle control such that lateral and longitudinal vehicle motion, monitoring of the driving environment\RemoveForShortVersion{ by detecting objects and events, identification of the vehicle's condition and its position in the environment} ;  
    
    \item[ii)] \textit{Tactical} components, which plan and execute vehicle maneuvers and prepare appropriate responses to incoming events, e.g. trajectory control, lane change\RemoveForShortVersion{, obstacle avoidance, emergency braking, visibility improvement by adapting lighting to environmental conditions};
    
    \item[iii)] \textit{Strategic} components, which determine the general itinerary according to the driver's preferences\RemoveForShortVersion{ and the traffic conditions, based on a maps database and a path planning algorithm}.
    
\end{enumerate}
    
Given the high complexity of the tasks to be automated, \textit{Machine Learning} approaches have become an essential element in the design of autonomous vehicles \cite{ma2020}. 
Machine learning is therefore used in the development of different classes of components:

\begin{enumerate}
    
    \item[i)] \textit{Operational} components are the most developed in the literature, and can be classified as follows: (1) \textit{mediated perception} approaches \RemoveForShortVersion{are mostly based on deep learning techniques} \cite{fagnant2015, john2015} \RemoveForShortVersion{and} aim to detect a wide variety of objects, such as obstacles, road signs\RemoveForShortVersion{, lanes and traffic lights} ; (2) \textit{direct perception} \cite{bojarski2016, bojarski2017} \RemoveForShortVersion{consists of end-to-end approaches, which} aim to directly manage vehicle controls \RemoveForShortVersion{(e.g., gas pedal, brake, steering wheel)} without explicitly dealing with location and mapping ; (3) \textit{localization} approaches \cite{alcantarilla2018, vishnukumar2017} \RemoveForShortVersion{aim to} characterize similarities and discrepancies between the environment \RemoveForShortVersion{observed from sensor data,} and a priori maps, \RemoveForShortVersion{in order} to \RemoveForShortVersion{accurately} locate the vehicle and \RemoveForShortVersion{identify potential} obstacles.     
    
    \item[ii)] \textit{Tactical} components are mostly developed to automate vehicle maneuvers using Machine Learning techniques, such as : (1) advanced scenarios of \textit{automated parking}\RemoveForShortVersion{, as free space recognition, pedestrian detection during the operation} \cite{heimberger2017} ; (2) \textit{car-following} improvement by predicting the trajectories of other human-drivers \cite{gong2018}\RemoveForShortVersion{, taking into account road conditions to increase safety} \cite{li2018} ; (3) \textit{trajectory planning} including obstacle avoidance [27], self-driving in urban environment \cite{sales2014} \RemoveForShortVersion{and at high speed \cite{al2002}, sliding control improvement \cite{akermi2020}}.
    
\end{enumerate}

\RemoveForShortVersion{An autonomous vehicle is an extremely complex system, which must react safely, and in real time, to the vagaries of its environment. 
The decisions made by the Machine Learning based components (presented above) are definitely \textit{time-sensitive}. In this case, reaching a good compromise between \textit{earliness} and \textit{accuracy} of these decisions is critical. 
On the one hand, early detection of an obstacle facilitates the planning of a safe evasive trajectory and perceived as such by the passengers. On the other hand, false positives can be generated by too early and not enough accurate detections, causing unnecessary or even dangerous trajectory changes.
}

%\medskip
Considering the \textit{earliness vs. accuracy} compromise is an emerging and important issue in research for autonomous vehicles. 
In particular, \textit{cooperative perception} \cite{kim2015} has been developed to extend the perceptual range of a connected autonomous vehicle, by sharing real-time information with other surrounding vehicles.  
\RemoveForShortVersion{
In this scenario, the vehicles' situational awareness is improved, allowing for smoother and safer maneuvers, such as early lane changes or early emergency braking.}
In some ways, cooperative perception improves both the \textit{earliness} and \textit{accuracy} of decisions by extending the vehicles' perceptual capability. 

\Alexis{TODO : parler de la cybersécurité dans une version longue??}

%\medskip
The development of the ML-EDM field could make it easier to design fully autonomous vehicles.  Indeed, the ability of these approaches to make \textit{non-myopic} decisions is an advantage to make self-driving more fluid and safe. 
\RemoveForShortVersion{
An experienced driver is capable of anticipating what is going to happen on the road: he is able to build a mental picture of the situations that are likely to occur in a few seconds.} 
A non-myopic ML-EDM approach would be able to identify probable continuations of an observed situation on the road, based on related situations encountered in the training data and their continuations\RemoveForShortVersion{ (e.g., if a balloon crosses the road, it is probable that a child will run behind)}. 

%\medskip
Most of the challenges presented in this article are relevant for the autonomous vehicle. Indeed, training data \RemoveForShortVersion{is very complex in this case, and} consists of multiple sources: video, radar\RemoveForShortVersion{, lidar, sensors} etc. In addition, training data may change over time, for example with the arrival of a new types of sensors on a next generation of car. It is therefore important to develop ML-EDM approaches which are \textit{agnostic} to data types (see challenge \#4). \RemoveForShortVersion{In the case of self-driving,} Ground truth contained in the training data includes both the actions to be performed \RemoveForShortVersion{(e.g., emergency braking)} and the timing of these actions \RemoveForShortVersion{(e.g., triggering this action 3 seconds before the potential impact)}. It would be very useful \RemoveForShortVersion{and probably possible} to develop ML-EDM methods which learn the evolution of \textit{decision costs} over time\RemoveForShortVersion{, underlying self-driving} (see challenge \#10). In \RemoveForShortVersion{the case of} autonomous vehicles, sensor data is continuously observed, so it is essential to design \textit{online} ML-EDM approaches (see challenges \#5 and \#6), and driving actions must definitely be \textit{revocable} (see challenges \#8 and \#9). \RemoveForShortVersion{For example, an emergency brake must not be carried out to the end, if there are finally no obstacles in the way.}

%-------------------------------------------------------------------
%-------------------------------------------------------------------
\section{Conclusion and perspectives}
\label{conclusion}
%-------------------------------------------------------------------
%-------------------------------------------------------------------

More and more applications require to make time-constrained decisions. 
On the one hand, an \textit{early} decision is based on partially observed data, leaving time before the deadline which allows for a proper response by performing the right actions. 
On the other hand, a \textit{late} decision based on nearly complete data tends to be more accurate, but leaves insufficient time to take appropriate action before the deadline. 
This compromise between the \textit{earliness} and the \textit{accuracy} of decisions has been particularly studied in the field of Early Time Series Classification (ECTS). 
In this paper a more general problem is introduced, called Machine Learning based Early Decision Making (ML-EDM), which consists in optimizing the decision times of models in a wide range of settings where data is collected over time.

This position paper aims to define the field of ML-EDM, and proposes ten challenges to the scientific community to further research in this area. 
In particular, ML-EDM has been defined and positioned with respect to related fields, such as machine learning and reinforcement learning. 
Three challenges have been presented in relation to the learning task at hand: extending ML-EDM to unsupervised learning (challenge \#1), to regression tasks (challenge \#2) and to weakly-supervised learning (challenge \#3).
The development of data type agnostic ML-EDM approaches has been singled out as an important direction of research to extend the domains of application, yielding challenge \#4.
Extending ML-EDM to the online scenario has also been recognized as important too which raises three challenges (challenge \#5, \#6 and \#7). 
Being able to revoke decisions properly is significant as well in many applications and raises two challenges (\#8 and \#9).
The origin of the different costs involved in the optimization of decision times has been discussed, leading to a last challenge (challenge \#10) to extend the ML-EDM problem to cases where these costs vary over time. 
Finally, a range of application areas for which ML-EMD could lead to significant progress in the near future have been described\RemoveForShortVersion{, such as anomaly detection, predictive maintenance, patient health monitoring, self-driving vehicles}.  

The overall objective of this position paper is to define a new field of investigation, and to propose research avenues in order to generate interest from the scientific community. 

\bibliographystyle{IEEEtran}
\bibliography{main}

%\begin{IEEEbiography}{Alexis Bondu}
%\end{IEEEbiography}

\section{Appendix}

\subsection{In practice, how to define the loss function ?}
\label{appendix1}

\begin{figure}[htbp!]
\centering
\includegraphics[width=0.7\linewidth]{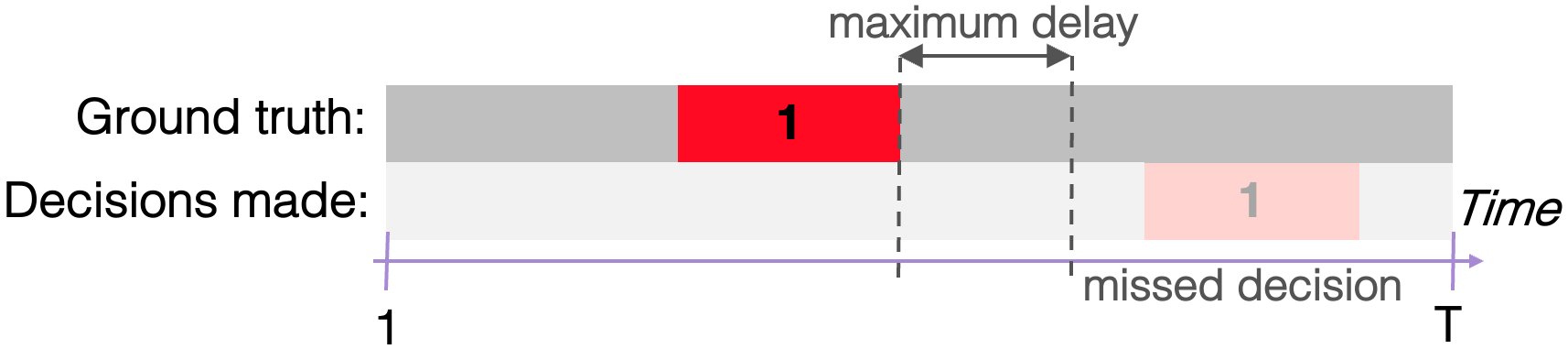}
%\vspace{-4mm}
\caption{Maximum delay after which a decision is considered as missed.}
\label{mapping-rule-max-delay}
\end{figure}

\begin{figure}[htbp!]
\centering
\includegraphics[width=0.7\linewidth]{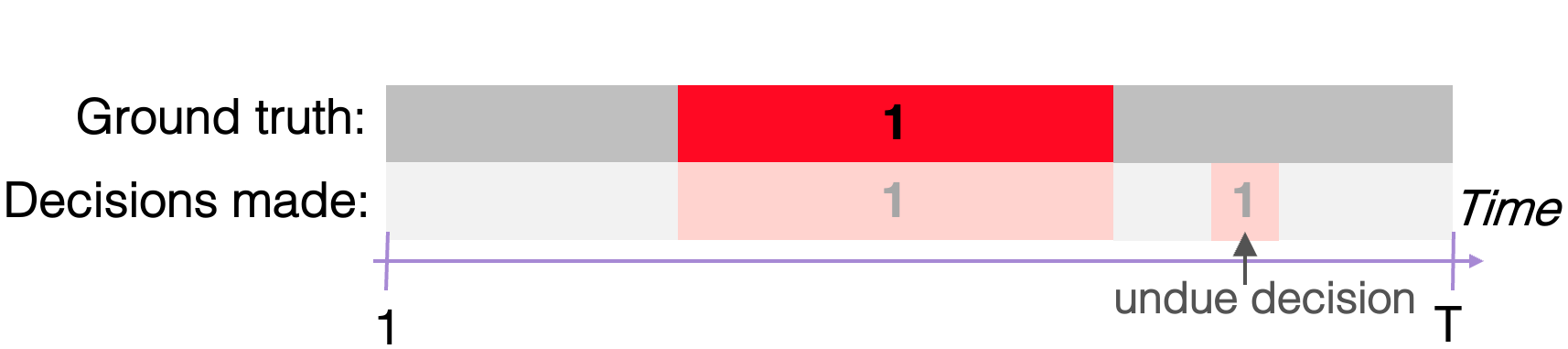}
%\vspace{-4mm}
\caption{A decision is undue if no true decision exists in the time interval.}
\label{mapping-rule-undue}
\end{figure}

\begin{figure}[htbp!]
\centering
\includegraphics[width=0.7\linewidth]{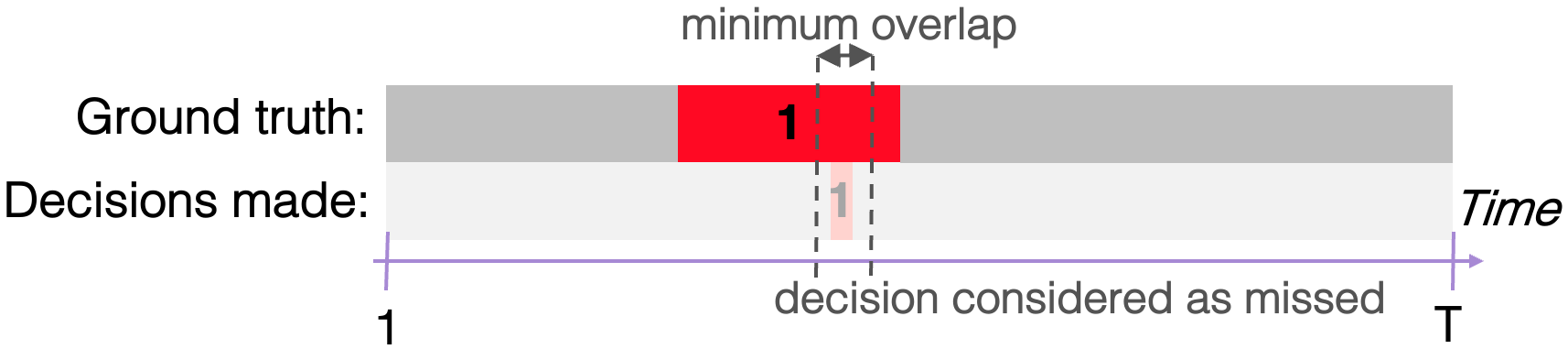}
%\vspace{-4mm}
\caption{Minimum overlap to consider that a decision is not missed.}
\label{mapping-rule-min-overlap}
\end{figure}

The loss function $\LL$ in Equation \ref{eq:double-loss-function} can be expressed in many different ways, depending on the application considered. 
In practice, \textit{mapping rules} need to be defined to match the decisions made to the true ones. In Equation \ref{eq:double-loss-function}, the purpose is to map the indices $i'$ to $i$, considering that the number of decisions made may be different than it should be ($\hat{k}_{\mathbf x} \neq k_{\mathbf x}$). 
In particular, these rules address the following questions: (\textit{i}) how long should we wait before considering that a true decision has been missed? (see a rule example in Figure \ref{mapping-rule-max-delay}) (\textit{ii}) when the number of decisions made is too large, how to identify the undue decisions? (e.g. Figure \ref{mapping-rule-undue}) (\textit{iii}) what is the minimum time overlap between a decision made and the corresponding true one? (e.g. Figure \ref{mapping-rule-min-overlap}) And of course, these mapping rules are specific to each application.

%\vspace{-3mm}

%\vspace{-3mm}

%\vspace{-3mm}

\begin{figure}[htbp!]
\centering
\includegraphics[width=0.7\linewidth]{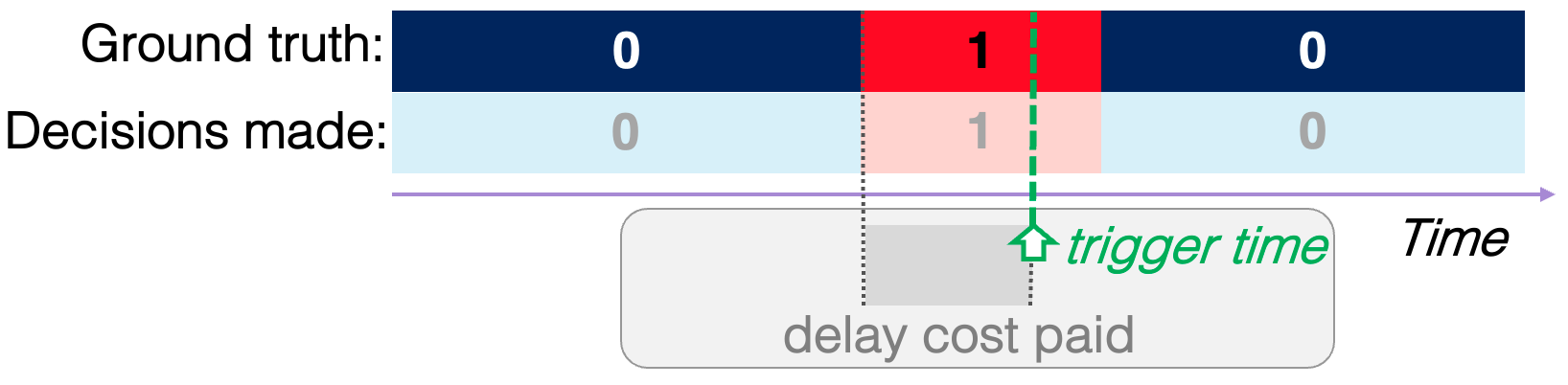}
%\vspace{-3mm}
\caption{Example of paid delay cost.}
\label{loss-delay}
\end{figure}

\begin{figure}[htbp!]
\centering
\includegraphics[width=0.7\linewidth]{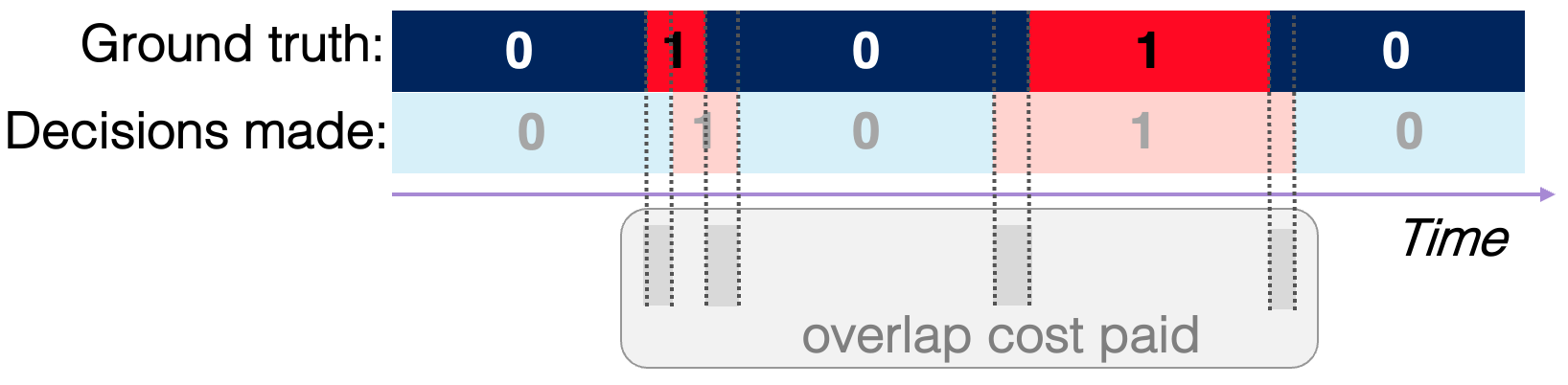}
%\vspace{-3mm}
\caption{Example of paid overlap cost.}
\label{loss-overlap}
\end{figure}

\begin{figure}[htbp!]
\centering
\includegraphics[width=0.7\linewidth]{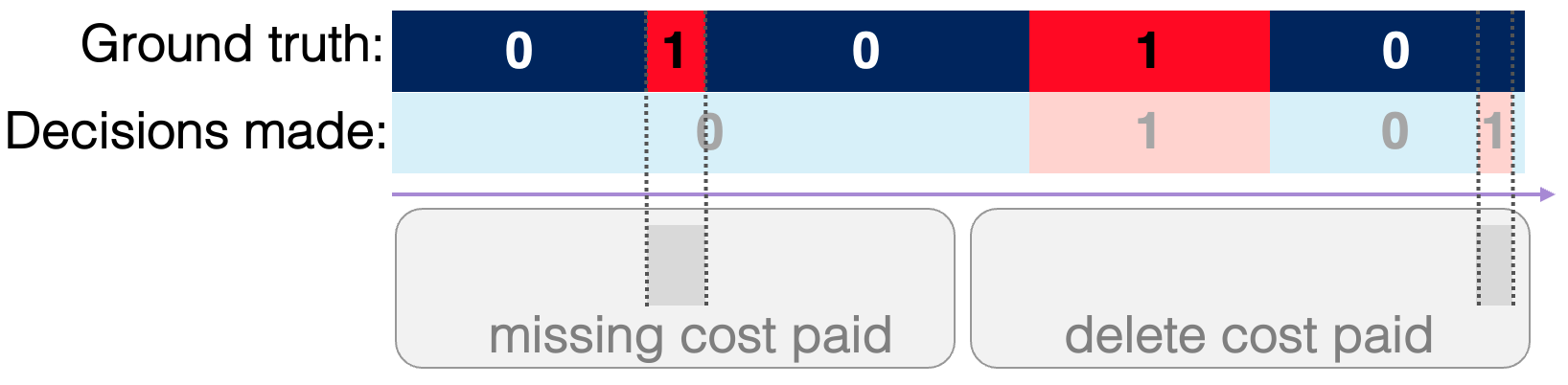}
%\vspace{-3mm}
\caption{Example of a missing decision and an extra undue one.}
\label{loss-missing-delete}
\end{figure}

In its general form, the loss function $\LL$ should involve several decision costs mentioned below. Their origin is further detailed in Section \ref{decision_costs}. For the moment, let us consider that the following costs are fixed, deterministic,  and given as input to an ML-EDM approach:
%\textcolor{gray}{
\begin{itemize}
    \item a prediction cost $\mathcal{L}_{prediction}$,  
    \item a delay cost $\mathcal{L}_{delay}$, 
    \item a time overlap cost $\mathcal{L}_{overlap}$, 
    \item a cost of missing the decision $\mathcal{L}_{missing}$, 
    \item a cost of an extra and undue decision $\mathcal{L}_{delete}$. 
    
\end{itemize}
%}
\smallskip
\noindent
As in the ECTS problem, the \textit{prediction} cost $\mathcal{L}_{prediction}$ accounts for a potentially bad prediction and it can be expressed as a cost matrix.
The delay cost $\mathcal{L}_{delay}$ depends on the trigger time $\hat{t}_{i'}$ and the time period associated to the i-th true decision $[s_i,e_i]$. Figure \ref{loss-delay} gives an example where a delay cost is paid since the triggering time (see the green vertical line) is located after the beginning of the period associated with the decision.   
The \textit{overlap} cost $\mathcal{L}_{overlap}$ accounts that the predicted periods $\{(\hat{s}_{i'}, \hat{e}_{i'})\}_{i'=1}^{\hat{k}_{\mathbf x}}$ 
might not coincide temporally with the periods of the true decisions $\{ {{(s_i,e_i)}}\}_{i=1}^{k_{\mathbf x}}$. 
For instance in Figure \ref{loss-overlap}, the decisions made (shown in the second line) are out of sync with the truth decisions (see the first line), which results in four overlapping periods. The interested reader may refer to \cite{tatbul2018} which addresses the evaluation of models by considering such temporal overlap.
Finally, the costs of missing a decision $\mathcal{L}_{missing}$ and making an additional undue one $\mathcal{L}_{delete}$ account that the number of decisions made can be different than it should be ($\hat{k}_{\mathbf x} \neq k_{\mathbf x}$). Figure \ref{loss-missing-delete} shows a situation where the first anomaly (represented by the class 1) is not detected, incurring a missing cost, and where a false detection occurs at the end leading to a delete cost. 

%-------------------------------------------------------------------
\subsection{In practice, how to evaluate a ML-EDM approaches?} 

\Youssef{TODO : étendre cette section ?? dans les cas où la décision prise modifie les donnée obervée, ex; en maintenance prédictive, on répare une machine qui ne serait pas tombé en panne, cf donnée sensurées.}

In some applications, decision costs are available as prior knowledge. It is the case for instance in \cite{khoshnevisan2021}, where the objective is to detect as early as possible patients suffering from septic shock, and where the cost of delaying decisions is perfectly known. When available, \textit{decision costs} are of great help in evaluating ML-EDM approaches. 
Indeed, each decision made can be evaluated by the amount of costs actually incurred, considering: (\textit{i}) the triggering moment ; (\textit{ii}) the ground truth ; (\textit{iii}) and the value of the decision costs (i.e. $\mathcal{L}_{delay}$, $\mathcal{L}_{prediction}$, and $\mathcal{L}_{revoke}$).  

In the particular case of ECTS problem, a cost-based evaluation criterion is proposed in \cite{achenchabe2021MLj} which simply corresponds to the \textit{empirical risk} calculated on a set of test individuals (see Equation \ref{risk_emp_8}).
In cases where decision costs are unavailable, a \textit{multi-criteria} evaluation can be considered to take into account the different costs. For instance, in \cite{mori2019early}, the \textit{earliness} and \textit{accuracy} of decisions are evaluated separately, and the Pareto optimal front is made up of the dominant approaches considering both criteria.

In the more general case of ML-EDM where multiple early decisions have to be made, a cost-based evaluation requires more prior knowledge. Indeed, mapping rules would have to be defined in order to match the triggered decisions with the true ones, the cost of overlap $\mathcal{L}_{overlap}$ between predicted and true time periods, as well as the costs of missing a decision $\mathcal{L}_{missing}$ and triggering an undue one $\mathcal{L}_{delete}$ (see Equation \ref{eq:double-loss-function} in Section \ref{definition-ML-EDM} and Figures \ref{mapping-rule-max-delay} to \ref{loss-missing-delete}).  

% CHANGE TEMPLATE

%\begin{IEEEbiography}[{\includegraphics[width=1in,height=1.25in,clip,keepaspectratio]{./figures/alexis.jpeg}}]{Alexis Bondu}
%\end{IEEEbiography}

%\begin{IEEEbiography}[{\includegraphics[width=1in,height=1.25in,clip,keepaspectratio]{./figures/youssef.jpeg}}]{Youssef Achenchabe}
%\end{IEEEbiography}

%\begin{IEEEbiography}[{\includegraphics[width=1in,height=1.25in,clip,keepaspectratio]{./figures/albert.jpg}}]{Albert Bifet}
%\end{IEEEbiography}

%\begin{IEEEbiography}[{\includegraphics[width=1in,height=1.25in,clip,keepaspectratio]{./figures/fabrice.jpg}}]{Fabrice Cl\'erot}
%\end{IEEEbiography}

%\begin{IEEEbiography}[{\includegraphics[width=1in,height=1.25in,clip,keepaspectratio]{./figures/antoine.jpg}}]{Antoine Cornu\'ejols}
%\end{IEEEbiography}

%\begin{IEEEbiography}[{\includegraphics[width=1in,height=1.25in,clip,keepaspectratio]{./figures/joao.jpg}}]{Joao Gama}
%\end{IEEEbiography}

%\begin{IEEEbiography}[{\includegraphics[width=1in,height=1.25in,clip,keepaspectratio]{./figures/georges.jpg}}]{Georges H\'ebrail}
%\end{IEEEbiography}

%\begin{IEEEbiography}[{\includegraphics[width=1in,height=1.25in,clip,keepaspectratio]{./figures/vincent.jpeg}}]{Vincent Lemaire}
%\end{IEEEbiography}

%\begin{IEEEbiography}[{\includegraphics[width=1in,height=1.25in,clip,keepaspectratio]{./figures/pierre-francois.jpg}}]{Pierre-François Marteau}
%\end{IEEEbiography}

\end{document}